\documentclass[twoside]{article}

\usepackage[accepted]{aistats2022}

\usepackage[round]{natbib}
\usepackage{my_style}
\usepackage{my_notation}
\graphicspath{{content/images/}}
\usepackage{caption}
\usepackage{subcaption}

\begin{document}

\twocolumn[

\aistatstitle{Regret, stability \& fairness in matching markets with bandit learners}

\aistatsauthor{ Sarah H. Cen \And Devavrat Shah }

\aistatsaddress{ Massachusetts Institute of Technology  \And Massachusetts Institute of Technology }
]

\begin{abstract}
	Making an informed decision---for example, when choosing a career or housing---requires knowledge about the available options. Such knowledge is generally acquired through costly trial and error, but this learning process can be disrupted by competition. In this work, we study how competition affects the long-term outcomes of individuals as they learn. We build on a line of work that models this setting as a two-sided matching market with bandit learners. A recent result in this area states that it is impossible to simultaneously guarantee two natural desiderata: stability and low optimal regret for all agents. Resource-allocating platforms can point to this result as a justification for assigning good long-term outcomes to some agents and poor ones to others. We show that this impossibility need not hold true. In particular, by modeling two additional components of competition---namely, costs and transfers---we prove that it is possible to simultaneously guarantee four desiderata: stability, low optimal regret, fairness in the distribution of regret, and high social welfare.
\end{abstract}

\section{INTRODUCTION}\label{sec:intro}

When individuals compete for resources, 
it is often assumed that every individual has access to knowledge 
that allows them to determine which resources they prefer. 
For example, 
consider choosing a primary care physician or selecting a career path. 
Some individuals may have insider knowledge about physicians or careers, 
e.g., due to previous experiences or advice relayed by friends. 
Others are left in the dark. 
Those in the dark can only obtain knowledge through costly trial and error, 
which can be further frustrated by the presence of competition. %
For example, if members of a group are repeatedly denied opportunities in academia due to competition,
they may never obtain the requisite knowledge to make an informed decision about pursuing a career in academia. 

In this work, we are interested in the interplay between \emph{learning} and \emph{competition}. 
Specifically, we are motivated by the question: 
How does competition affect an individual's ability to make informed decisions
and ultimately the individual's long-term outcomes?

To address this question, we study
the \emph{two-sided, competitive matching market}. 
In this setting, there are two types of agents.
For linguistic convenience, 
we call agents on one side of the market \emph{users} and those on the other \emph{providers}. 
Each user has a set of preferences over providers, and vice versa. %
In a matching market, 
users and providers are paired one-to-one in an assignment is known as the \emph{matching}. 
Since a user can be paired to at most one provider at a given time, and vice versa, 
there is \emph{competition} between agents whose preferences overlap. 
Such markets are ubiquitous and include
sellers competing to show advertisements to social media users, 
students competing for internships, 
residents competing for housing, 
and more. 

In order to study how knowledge affects the outcomes of individuals in competition, 
we extend this problem to include a \emph{learning} component, as described next.

	\begin{figure*}[t!]
	\centering	
	\includegraphics[width=\textwidth]{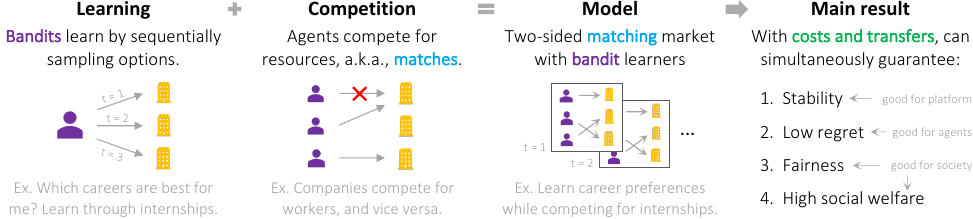}
	\caption{We study learning under competition using the two-sided matching market with bandit learners. As our main result, we show that, with costs and transfers, it is possible to simultaneously guarantee four desiderata.}
	\label{fig:contributions}
\end{figure*}

\textbf{Knowledge in matching}.
In the standard centralized matching problem, as posed by \citet{gale1962college}, 
every user knows their true preferences over providers \emph{a priori}, and vice versa. 
Agents then report their preferences to a central platform, which assigns a matching. 
A matching $M$ is \emph{stable} if there is no user-provider pair such that they prefer to be matched to one another rather than to their matches under $M$. 
Intuitively, when matchings are stable, 
agents trust the platform in the sense that
there is no incentive for them to defect. 
While stability is desireable for the platform,
it does not necessarily reflect overall performance. 
Indeed, as studied by \citet{axtell2008high}, 
stability can come at the expense of other objectives, 
such as  \emph{high social welfare} or \emph{fairness}. 

Recent works study an extension of the matching problem in which agents are \emph{not assumed to be knowledgeable}. %
They build on the insight that, 
contrary to the core assumption of the standard matching problem, 
users and providers rarely know their true preferences \emph{a priori}. 
Rather, the agents learn their preferences through experiences---specifically, by being matched. 
For example, consider the internship market in which workers present employers with qualifications, and employers present workers 
with career options. 
In this setting, matching workers to internships allows workers to learn their preferences over careers and employers to learn their 
preferences over worker qualifications.

\textbf{Blending matching and bandit learning}.
While the standard matching problem captures agents competing for  resources, 
a more sophisticated model is needed to 
model how agents learn through trial and error. 
Conveniently, 
the task of \emph{learning one's preferences from experiences} is at the core of the multi-armed bandit (MAB) problem \citep{cesa2006prediction, bubeck2012regret}.

In this way, one can study matching under learning by blending the matching and MAB problems.
This is precisely what \citet{liu2020competing} do in a recent work on the {centralized two-sided matching problem with bandit learners}. 
Remarkably, \citet{liu2020competing} find that, although agents may not know their true preferences \emph{a priori},
it is possible to assign matchings that are
stable at every time step
\emph{and}
permit agents to learn enough
so that the system converges to matchings that are stable under the agents' 
\emph{true} (unknown) preferences. 

\textbf{Our focus}.
\citet{liu2020competing} also surface an impossibility result:
a matching platform cannot guarantee low \emph{optimal} regret---a measure of long-term happiness---for all agents without sacrificing stability. 

This impossibility result is the focus of our work. 
Consider online platforms (e.g., Facebook), 
which are generally concerned with the retention of agents, such as their users, content sources, and advertisers.
Recall that a lack of stability implies that agents are better off when they leave the platform. 
Therefore, online platforms are typically permitted to prioritize stability because it is tied to their financial survival. 
The impossibility result above gives platforms a justification for favoring some agents over others. 
In other words, 
when a platform chooses matches that assign good long-term outcomes to some while assigning poor ones to others,
the platform can argue that it could not have done any better by pointing to the impossibility result.
\emph{But does this impossibility result hold in general or are there ways to overcome it?}

In this work, we show that the impossibility result is an artifact of the model posed by  \citet{liu2020competing}, 
i.e., that stability and low regret need not be in tension.
By adding a layer of complexity, 
we find that it is possible to \emph{simultaneously guarantee stability, low regret, fairness, and high social welfare}. 

\textbf{Matching with costs and transfers}.
Our adjustment is the modeling of {costs} and {transfers}.
Although this adjustment is minor, 
it can overturn the impossibility result and align the platform's and agents' interests.
We provide several examples showing how costs and transfers might arise in real-world settings.
Intuitively, 
competition can give rise to payments---which are equivalent to \emph{transfers} between agents---or demand additional effort and fees from agents---which are equivalent to \emph{costs}.

Our contributions are visualized in Fig. \ref{fig:contributions} and summarized as follows. 
We formulate the centralized, two-sided matching problem with bandit learners under  costs and transfers in Section \ref{sec:problem_statement}.
In Section \ref{sec:results}, 
we show that, without costs or transfers, it is impossible to guarantee stability alongside low regret, fairness, or high social welfare (Proposition \ref{prop:no_P_no_T}). 
We then show that, with costs and transfers, it is possible to simultaneously guarantee all four desiderata (Theorem \ref{thm:possibility_result}).
In Section \ref{sec:examples}, 
we provide intuition via several examples.

	\section{RELATED WORK}\label{sec:related_work}
	
	\noindent \textbf{Two-sided matching}.
	In the two-sided matching problem, there are two distinct populations
	that wish 
	to be matched one-to-one.
	Examples of two-sided matchings range from how individuals find partners or how workers fill job openings to how platforms match content to users or how doctors are assigned to patients. 
	One of the challenges is finding a \emph{stable} matching:
	a matching under which no pair of agents is incentivized to defect and pair off together. 
	
	In a foundational work, 
	\citet{gale1962college} show that there exists at least one stable matching and provide an algorithm---known as the Gale-Shapley (GS) or deferred acceptance algorithm---that finds such a matching \citep{knuth1997stable,roth2008deferred}.
	In the footsteps of \citet{gale1962college}, one line of research studies \emph{centralized} matchings, in which agents' preferences are reported to a central platform that assigns the matches. 
	Another line of work studies decentralized (or distributed) matching with no coordination between agents \citep{roth1990random,echenique2012experimental}.
	
	There are many closely related problems, such as the 
	the generalized assignment \citep{kuhn1955hungarian,ross1975branch,pentico2007assignment};
	housing allocation \citep{shapley1974cores}; 
	college admissions or hospital-residents \citep{gale1962college,roth1984evolution};  
	stable roommates  \citep{irving1985efficient}; and 
	generalized stable allocation 
	\citep{dean2009generalized} problems.
	
	\textbf{Multi-armed bandit (MAB) problem}.
	In the MAB problem  \citep{bubeck2012regret,lattimore2020bandit}, an agent (the bandit) sequentially chooses between $K$ options (or arms).
	At each time step, the agent selects an arm and receives a noisy reward in return. 
	The agent's goal is to maximize its cumulative reward over $T$ time steps or, equivalently, to minimize \emph{regret}. 
	However, the agent does not know the expected rewards of each arm \emph{a priori}. 
	To minimize regret, 
	the bandit learner must balance its short-term desire to select arms with high, known rewards against its need to learn the expected reward of other arms.
	This balance---also known as the exploitation-exploration trade-off \citep{wilson2014humans,schnabel2018short,wilson2020balancing}---is a focal point of reinforcement learning  \citep{sutton2018reinforcement,wiering2012reinforcement}. 

	The goal of the MAB problem is to design a selection policy such that the agent incurs low regret. 
	\citet{lai1985asymptotically} showed that regret must grow $\Omega(\log(T))$, which led to research on policies that guarantee $O(\log(T))$  regret
	\citep{thompson1933likelihood,bubeck2012regret,lattimore2020bandit}. 
	One of the most popular policies is known as the \emph{upper confidence bound (UCB)} strategy \citep{auer2002finite}, which is a deterministic decision-making strategy that achieves low, i.e.,  $O(\log(T))$, regret.

	\noindent \textbf{Matching markets with bandit learners}. 
	Sequential decision-making under uncertainty and competition is a problem of long-standing interest. 
	Recently, such settings have been modeled as a combined matching and MAB problem  \citep{das2005two,liu2020competing}.
	While  \citet{das2005two} introduce the combined problem, they study it
	under the strict assumption that agents on one side of the market have identical preferences.  
	In this work, we study the setting
	presented by \citet{liu2020competing}.
	
	In the centralized matching market with bandit learners \citep{liu2020competing},
	users and providers are on opposite sides of the market. 
	As in the matching problem, users compete for matches with providers, and vice versa. 
	As in the MAB problem, agents do not know their true preferences \emph{a priori}.
	Instead, agents learn their preferences from previous matches. 
	At each time step, each user computes its UCBs for all providers, and vice versa, then reports them to the central platform, which runs the GS algorithm to assign a stable matching at each time step. 
	
	There are related works on how competition affects learning \citep{mansour2018competing,aridor2020competing},  bandits with collisions \citep{liu2010distributed,kalathil2014decentralized,bubeck2020non}, 
	and coordinated resource allocation \citep{avner2019multi}.
	Though related, these settings are one-sided (i.e., only one side of the market has preferences), and they do not consider costs or transfers.
	There are also recent works that study other aspects of the matching problem with bandit learners, such as decentralization, matching robustness, and information exchange \citep{chawla2020gossiping,boursier2020selfish,sankararaman2020dominate,vial2020robust,liu2020bandit}. 
	A recent work by \citet{jagadeesan2021learning} study a closely related problem of learning matchings 
	under transferable utilities. 
	These authors consider a different (slightly stronger) notion of stability and they allow transfer rules that vary each time step.

	\noindent \textbf{Costs and transfers}. 
	Recall that, in the MAB problem,
	an agent's regret is low if it is $O(\log(T))$. 
	\citet{liu2020competing} show that, in the centralized two-sided matching problem with bandit learners, it  is possible to guarantee stability and $O(\log(T))$ \emph{pessimal} regret for all agents. 
	However, there is no guarantee that an agent's \emph{optimal} regret is $O(\log(T))$.
	We propose this impossibility result can be resolved by adding a layer of complexity not present in the previous work by \cite{liu2020competing}: \emph{costs} and \emph{transfers}.
	Intuitively, due to competition in matching markets, it is natural for transfers (e.g., payments) to arise 
	or for agents to internalize costs (e.g., fatigue).

	Incorporating transfers in matching is not new. 
	Several works on the standard  two-sided matching problem  allow transfers between agents \citep{shapley1971assignment,becker1973theory}. 
	They show that transfers are important to balancing supply and demand, allowing markets to clear
	\citep{shapley1971assignment,kelso1982job}.
	It has also been suggested that transfers (e.g., prices) and costs (e.g., search frictions) can explain discrepancies between predicted and observed matching behaviors \citep{hitsch2010matching}.

	\noindent \textbf{Stability, fairness, and social welfare}. 
	Early works on matchings study stability \citep{gale1962college}
	while later works consider other objectives, such as 
	high social welfare \citep{axtell2008high}, 
	fairness \citep{masarani1989existence,klaus2006procedurally,suhr2019two},
	envy-freeness \citep{wu2018lattice},
	and strategy-proofness \citep{roth1982economics,ashlagi2018stable}. 
	While important, stability is a local objective and does not necessarily reflect global performance. 
	For example,
	stability can come at the expense of social welfare (the sum of all agents' payoffs) 
	and fairness (the balanced distribution of regret across agents)  \citep{axtell2008high}. 
	In this work, we consider stability, fairness, and social welfare. 
	We do not study envy-freeness and strategy-proofness. 
	As our main contribution, we show that it is possible to simultaneously guarantee stability, fairness, and high social welfare as well as low regret.

\section{PROBLEM STATEMENT}\label{sec:problem_statement}

In this section, 
we formalize the centralized matching problem with bandit learners. 
The setup is similar to that given by \citet{liu2020competing}. 
Our main modification is the incorporation of costs and transfers, 
and our definitions are adjusted accordingly.

\subsection{Setup}\label{sec:setup}

	Suppose  that there are two types of agents who sit on opposite sides of the market. 
	For linguistic convenience, we call the agents on one side \emph{users} and those on the other \emph{providers}.
	We denote the sets of users and providers by $\cU = \{ u_1, u_2, \hdots, u_N \}$ and $\cP = \{p_1, p_2, \hdots, p_L \}$, respectively, where $\cU \cap \cP = \emptyset$.
	Let $\cA =  \cU \cup \cP$ denote the set of all agents and $\cA^+ = \cA \cup \{\emptyset\}$, where $\emptyset$ denotes no agent. 
	Without loss of generality, let $N \geq L$. 

	Each user has a set of preferences over providers, and vice versa.
	These \emph{true} preferences are unknown \emph{a priori} and must be learned with time. 
	For example, the true preference of an applicant for a job could capture how enjoyable the applicant would find the job if given the job, and the true preference of an advertiser for a Facebook user would be how profitable that user is to the advertiser if matched to it. Here, the ``agents'' are the applicants and jobs in the first example, and the advertisers and Facebook users in the second example. 
	
	Let $\mu(a_1, a_2)$ denote agent $a_1$'s true (unknown) preference for agent $a_2$, where $\mu: \cA \times \cA^+ \rightarrow \bbR_{\geq 0}$, and $a_1$ and $a_2$ cannot both belong to $\cU$ or to $\cP$. 
	We assume that there are no ties, i.e., $a' \neq a'' \implies \mu(a, a') \neq \mu(a, a'')$ for all $a \in \cA$. 
	We also assume that 
	$\mu(a,a') + \mu(a',a) \neq \mu(a''',a'') + \mu(a'',a''')$ unless $(a, a') = (a'', a''')$ or $(a, a') = (a'', a''')$.
	This assumption can be relaxed, as we discuss in the Appendix. 

	\textbf{Matching}.
	Let $T \in \bbN_{>0}$ be the time horizon. At every time step $t \in [T]$, a platform matches users and providers based on the agents' preferences $H_t$ at time $t$. 
	Let $\ccM(a ; H_t )$ denote the agent to which $a$ is matched at time $t$, where $\ccM: \cA \rightarrow \cA^+$ denotes the \emph{matching}, and $\ccM(a ; H_t ) = \emptyset$ implies that agent $a$ is unmatched at time $t$. 
	Without loss of generality, we assume that agents always prefer to be matched than to be unmatched. 
	A \emph{feasible} matching is one in which each user is matched to at most one provider, and all providers are matched to exactly one user. 
	Let $\cW$ be the set of all feasible matchings between users and providers.
	
	An agent $a$ learns how compatible they are with agent $a'$ through trial and error. 
	Formally, if $a$ is matched to $a'$ at time $t$, then $a$ receives a reward $X_t(a,a')  \iid \cD(a , a')$,
	where $\cD(a,a')$ is a sub-Gaussian distribution with parameter $\sigma^2$ centered at $\mu(a,a')$.
	We adopt the convention that $\mu(a,\emptyset) = X_t(a,\emptyset) = 0$. 
	Each agent may also incur a cost and/or partake in a transfer. 
	Let $\cC(a, a' ; H_t)$ denote the \emph{cost} that $a$ would incur if matched to $a'$ at time $t$, where $\cC: \cA \times \cA^+ \rightarrow \bbR$.
	Let $\cT(a , a' ;  H_t)$ denote the \emph{transfer} that $a$ would receive from $a'$ if matched to $a'$ at time $t$, where $\cT: \cA \times \cA^+ \rightarrow \bbR$ and $\cT(a , a' ;  H_t) = - \cT(a' , a ;  H_t)$.  
	We use the convention that $\cC(a,\emptyset) = \cT(a, \emptyset) = 0$ for all $a \in \cA$. 
	
	Therefore, if agent $a$ is matched to $a'$  based on preferences $H_t$, then $a$'s \emph{observed payoff}  is 
	\begin{align*}
		U_t(a, a' ; H_t)= X_t(a , a') - \cC(a, a' ; H_t) + \cT(a ,a' ; H_t) .
	\end{align*}
	
	Each agent observes its own match, reward, cost, and transfer but not the private information of any other agent.
	Other than the matchings, costs, and transfers that it assigns, the platform only observes the preferences $H_t$ at time $t$. The platform cannot observe the agents'  other private information. 

	\noindent \textbf{Bandit learners}. 
	We refer to the agents as \emph{bandit learners} because their task---to learn one's preferences by sequentially sampling the options---mirrors the MAB problem.
	Let each agent $a \in \cA$ begin with $T_0(a,a')$ samples 
	$\{X^1(a,a') \, , \,  \hdots \, , \, X^{T_0(a,a')}(a,a') \}$,
	 where $X^\tau(a,a') \iid \cD(a,a')$ for all $\tau \in [T_0(a,a') ]$ and $T_0(a,a') = T_0(a',a)$.
	 Let $T_0(u, p) \, , \, T_0(p, u) > 0$ for all $u \in \cU$ and $p \in \cP$.
	 In words, agents have at least one noisy sample before matching begins.\footnote{
		This construction is equivalent (with small modifications) to cyclically matching users and providers in the first $N$ time steps, as often done in the MAB setting.	
}
	Let $T_{t}(a, a') = T_0(a,a') +  \sum_{\tau \in [t]} \mathbf{1}( \ccM(a ; H_{\tau}) = a' )$
	be the number of samples $a$ has for $a'$ up to time $t$ and $\hat{\mu}_t(a, a') = ( T_{t-1}(a, a') )^{-1} \big( \sum_{\tau \in [t-1]} \mathbf{1}( \ccM(a ; H_{\tau}) = a' ) X_{\tau}(a, a') + \sum_{\tau \in [T_0(a,a')]} X^{\tau}(a,a')  \big)$
	denote the empirical mean of the samples.
	
	Under the $\sigma^2$-sub-Gaussian assumption above,
	the $\alpha$\emph{-upper confidence bound (UCB)} of $a$ for $a'$ is:
	\begin{align*}
		\nu_t(a,a')  = \hat{\mu}_t (a,a') + \sqrt{ ( 2 \sigma^2 \alpha \log(t) ) / T_{t-1}(a, a') } ,
	\end{align*}
	where $\alpha \in \bbR \cap (2, \infty)$ 
	and $\nu_t(a,\emptyset) = 0$ for all 
	$a, a' \in \cA$. 
	We call $\nu_t(a, \cdot)$ the \emph{transient} preferences of agent $a$ at time $t$. 
	To see why, note that in the absence of competition and if $\cC = \cT = 0$, our setup is identical to the classical MAB problem, 
	and $\nu_t(a,a')$ captures $a$'s preference for being matched to (i.e., sampling) $a'$ at time $t$ under the UCB algorithm. 

	In this work, we let $H_t = \nu_t$, i.e., agents report their transient preferences $\nu_t$ to the platform at time $t$.\footnote{Letting $H_t = \nu_t$ is natural because an agent's preferences at a given time $t$ must balance exploration and exploitation, as captured by UCB preferences. Note that we could equivalently let $H_t = \hat{\mu}_t$, i.e., agents report their empirical estimates of $\mu$. Our results would still hold. The only modification would be that the platform computes $\nu_t$ from $\hat{\mu}_t$ based on the matching history.}
	
	\noindent \textbf{Summary of matching process}.
	In summary, the matching process with costs and transfers proceeds as follows. 
	At $t = 0$, the platform decides on 
	 $(\ccM, \cC, \cT)$, which is made known to all agents. 
	Then, at every time step $t \in [T]$, there are four stages:
	\begin{enumerate}[noitemsep]
		\item  \emph{Update}: Every agent $a$ updates estimates $\hat{\mu}_t(a,\cdot)$. 
		\item  \emph{Report}: Using $\hat{\mu}_t(a,\cdot)$, every agent $a$ reports its transient preferences $\nu_t(a,\cdot)$ to the platform.
		\item  \emph{Match and Observe}: Based on $\nu_t$, the platform matches users and providers according to $\ccM(\cdot \, ; \nu_t)$.
		Each agent $a$ observes its own match $\ccM(a; \nu_t)$ and stochastic reward $X_t(a , \ccM(a ; \nu_t))$. 
		\item \emph{Pay and transfer}: Each agent $a$ may incur the cost $\cC(a , \ccM(a ; \nu_t) ; \nu_t)$ and/or receive the transfer $\cT(a , \ccM(a ; \nu_t) ; \nu_t)$ from its match $\ccM(a ; \nu_t)$. 	
	\end{enumerate}

	As mentioned above, the platform can only observe the agents' transient preferences as well as the matches, costs, and transfers that it assigns to the agents. 
	An agent can only observe its own information, including its preferences, match, reward, cost, and transfer.

	\subsection{Desiderata}\label{sec:objectives}
	
	The platform's role is to design the matching mechanism $\ccM$, cost rule $\cC$, and transfer rule $\cT$.
	In this work, there are four desiderata of interest: \emph{stability}, \emph{regret}, \emph{fairness}, and \emph{social welfare}.

	We first introduce some notation. 
	We say that $V(a,a')$ is the payoff that agent $a$ receives if matched to agent $a'$
	under some payoff function  $V: \cA \times \cA^+ \rightarrow \bbR$.
	When we are concerned with a specific set of preferences $\psi : \cA \times \cA^+ \rightarrow \bbR$, 
	we use the following notation:
	\begin{align*}
		V(a , a';  \psi ) = \psi( a , a' ) - \cC( a, a';  \psi) + \cT(  a , a' ; \psi) ,
	\end{align*}
	for all $a \in \cA$ and $a' \in \cA^+$. 

	\noindent
	\emph{\textbf{(a) Stability}}. 
	A matching $\sM \in \cW$ is \emph{not} stable under $V$ if there exist
	$a, a' \in \cA$ such that $V(a , \sM(a)  ) < V(a, a'  )$
	 and $V(a', \sM(a') )  < V(a' , a )$.\footnote{In some works, this definition of stability is known as ``weak stability'' \citep{irving1994stable,manlove2002structure}. In this work, when we use ``stability'', we mean ``weak stability''.}
	In words, a matching is not stable when there exists a user-provider pair that, under payoffs $V$, would rather defect than participate in the matching. 
	Let $S(V) \subset \cW$ be the set of stable matchings under payoffs $V$. 
	\begin{definition}\label{def:defection_proof}
		$(\ccM , \cC , \cT)$ is \emph{stable} if and only if $\ccM(\cdot \, ; \nu_t) \in S(V(\cdot, \cdot \, ; \nu_t ))$ for all $t \in [T]$.
	\end{definition}

	 \noindent 
	 \emph{\textbf{(b) Low regret}}. 
		Regret measures the loss incurred by an agent as it learns.
		Let the \emph{optimal} and \emph{pessimal} matchings for agent $a$ at time $t$ be defined as:
		\begin{align*}
			\bcM^a_t &\in \arg \max_{\sM \in S(V(\cdot, \cdot \, ; \mu )) } \bbE_{X_t \sim \cD} \left[ U_t(a , \sM(a) ; \nu_t ) \right] ,
			\\
			\ucM_t^a  &\in  \arg \min_{\sM \in S(V(\cdot, \cdot \, ; \mu)) } \bbE_{X_t \sim \cD} \left[ U_t(a , \sM(a) ; \nu_t ) \right] ,
		\end{align*}
		respectively.
		For intuition, recall that a stable matching is not necessarily unique, i.e., $|S(V(\cdot, \cdot \, ; \mu)) | \geq 1$.
		$\bcM^a_t$ and $\ucM_t^a$ are both stable matchings under $V(\cdot, \cdot \, ; \mu)$
		that, respectively, yield the highest and lowest expected payoffs for agent $a$ at time $t$. 
		Let the \emph{optimal regret} $\bR(a ; \ccM)$ of agent $a$ under $(\ccM, \cC, \cT)$ be: %
		\begin{align*}
			 \sum_{t=1}^T 
			\bbE [
			U_t(a , \bcM^a_t(a) ; \nu_t ) - 	U_t(a  , \ccM(a ;  \nu_t ) ; \nu_t  )
			] .
		\end{align*}
		Let \emph{pessimal regret} be defined analogously, exchanging $\bcM^a_t$ for $\ucM^a_t$.\footnote{
			Note that we often suppress $\cC$ and $\cT$ in our notation, such as in $\bar{R}$, $V$, and $W_t$. 
		}
		Note that $\bR(a ; \ccM )  \geq \uR(a ; \ccM )$. 
		If there is a unique stable matching, i.e., $|S(V(\cdot, \cdot \, ; \mu))| = 1$, then $\bcM_t^a = \ucM_t^a$ and $\bR(a ; \ccM) = \uR(a ; \ccM)$ 
	for all $a \in \cA$ and $t \in [T]$.
	We define low regret as follows.

	\begin{definition}\label{def:low_regret}
		An agent incurs \emph{low regret} when 
	$\uR(a ; \ccM) \leq \bR(a ; \ccM) = O(\log(T))$.
	\end{definition}

	\noindent \emph{\textbf{(c) Fairness}}.
	We say that
	$(\ccM, \cC, \cT)$ is unfair if some agents incur low regret while others incur high regret.
	\begin{definition}\label{def:fairness}
		$(\ccM,\cC,\cT)$ is \emph{unfair} if there exists a pair of agents $(a,a')$ such that $\bar{R}(a ; \ccM) = O(\log(T))$  and $\bar{R}(a' ; \ccM) = \omega(\log(T))$. 
		Otherwise, it is fair. 
	\end{definition} 
	
	\noindent \emph{\textbf{(d) High social welfare}}.
	Social welfare is a utilitarian metric defined as the sum of all agents' payoffs.
	Formally, the \emph{social welfare} of $(\ccM, \cC, \cT)$ at time step $t$ is $W_t(\ccM) = \sum_{a \in \cA} V(a, \ccM(a ; \nu_t) ; \nu_t)$. 
	\begin{definition}\label{def:SW}
		$(\ccM, \cC, \cT)$ guarantees $\kappa$-\emph{high social welfare} at time $t$ if $W_t(\ccM ) \geq 
		\kappa 
		\max_{\ccM'} W_t(\ccM')$
		and $W_t(\ccM ) > 0$. 
		If there does not exist such a $\kappa > 0$ or $W_t(\ccM ) = 0$,
		then the social welfare at $t$ is low.
	\end{definition}

\section{MAIN RESULTS} \label{sec:results}

	In this section, 
	we give two results.
	Using our setup, we first re-state the impossibility result of \citet{liu2020competing}. 
	We then give our main result, 
	which says that,
	with costs and transfers,
	one can simultaneously guarantee
	stability, low regret, 
	fairness, and high social welfare.
	All proofs can be found in the Appendix.

	Before proceeding, 
	one tool we require is a classical algorithm known as the Gale-Shapley (GS) algorithm, 
	which we restate in Algorithm \ref{alg:GS_short}.
	Several well-known results about the GS algorithm are given in Appendix \ref{sec:app_GS}. 
	The most relevant result for the understanding of this work is that the GS algorithm
	 always returns a stable matching relative to the given payoff function.

		\begin{algorithm}[t]
		\SetAlgoLined
		\KwIn{
			Set of agents $\cA = \cU \cup \cP$, 
			where $\cU = \{u_1, u_2, \hdots, u_N \}$, 
			$\cP  = \{ p_1, p_2, \hdots, p_L \}$, $\cU \cap \cP = \emptyset$, and $N \geq L$. Payoff function $V: \cA \times \cA^+ \rightarrow \bbR$.
	}
		\KwOut{Matching $\sM \in S(V) \subset \cW$. }
		\BlankLine
		
		Initialize $\sM: \cA \rightarrow \cA^+$ s.t. $\sM(a) \leftarrow \emptyset \,\, \forall a \in \cA$\;

		\While{$\exists$ \text{unmatched provider} $p \in \cP : \sM(p) = \emptyset$} {
			Let $u$ denote $p$'s most preferred user to which $p$ has not yet proposed\;

			If 
			$u$ is unmatched, i.e.,
			 $\sM(u) = \emptyset$, 
			 then 
			match $u$ and $p$ such that:
			$\sM(u) \leftarrow p$ and $\sM(p) \leftarrow u$\;

		Otherwise, if $u$ prefers $p$ to current match, i.e., $V(u,p ) > V(u , \sM(u) )$, match $u$ and $p$ s.t.	$\sM'( \sM(u) ) \leftarrow \emptyset$ , 
		$\sM(u) \leftarrow p$, 
		and
		$\sM(p) \leftarrow u$\;
		
		}
		\caption{Gale-Shapley algorithm (with providers as proposers)} \label{alg:GS_short}
		\algorithmfootnote{
			The GS algorithm is given in detail in the Appendix.} 
	\end{algorithm}
	
	\textbf{Impossibility result}.
	We begin with an impossibility result
	that  is analogous to that given by \citet{liu2020competing} with minor modifications.  
	It states that, in the \emph{absence} of costs and transfers,
	all agents are guaranteed to have low \emph{pessimal} regret, 
	 but there is no guarantee that they have low \emph{optimal} regret,
	 which can grow $\Omega(T)$. 
	Let:
	\begin{align*}
	&	\Delta_{\max} (a) =   \max_{a_1,a_2 \in \cA^+} (\mu(a , a_1 ) 
	- \mu(a , a_2 ) ) ,
		\\
	&	\Delta_{\min} = \min_{a_1 \in \cA, a_2 \in \cA^+, a_3 \in \cA^+ \setminus a_2} |\mu(a_1, a_2) - \mu(a_1, a_3)| .
	\end{align*}
	
	\begin{proposition}  \label{prop:no_P_no_T}
		Suppose that there are no costs or transfers such that:
		$\cC(a_1,a_2 ; \psi) = 0$ and $\cT(a_1,a_2; \psi) = 0$ %
		for all $a_1 \in \cA$, $a_2 \in \cA^+$, and $\psi: \cA \times \cA^+ \rightarrow \bbR$.  
		If the GS algorithm is applied over $V(\cdot, \cdot \, ; \nu_t)$ at every $t \in [T]$, then the system is stable. 
		Moreover, 
		$
			\uR(a ; \ccM) \leq 2 N^2 L  \Delta_{\max} (a)   \left( \frac{8\sigma^2 \alpha \log(T)}{\Delta_{\min}^2} + \frac{\alpha}{\alpha - 2} \right)
		$
		for all $a \in \cA$. 
		However, under stability, it is not possible to guarantee fairness or high social welfare, and there exist settings $(\cA, \mu)$ for which $\bR(a ; \ccM) = \Omega(T)$ for at least one agent $a \in \cA$. 
	\end{proposition}

	Proposition \ref{prop:no_P_no_T} states that applying the GS algorithm at each time step ensures that every matching is stable and that every agent has low pessimal regret.
	However, the second half of Proposition \ref{prop:no_P_no_T} reveals an impossibility result:
	without costs or transfers, 
	it is not possible to guarantee low \emph{optimal} regret alongside stability. 
	In fact, there are settings in which the optimal regret of at least one agent grows $\Omega(T)$.
	Guaranteeing low pessimal regret reassures pessimistic agents  that they will at least reach their worst-case performance under a true stable matching in $O(\log(T))$ time steps. 
	However, because low optimal regret is not guaranteed,  any optimistic agent will be disappointed: it can take $\Omega(T)$ time steps to surpass their worst-case performance under a true stable matching. 
	
	\textbf{Incorporating costs and transfers}.
	Therefore, without  costs or transfers,
	the platform cannot simultaneously guarantee stability and fairness or high social welfare. 
	This negative result begs the question: Is it possible to do better and guarantee low optimal regret, fairness, or high social welfare alongside stability?
	It turns out that
	the answer is yes.

		\begin{theorem}  \label{thm:possibility_result}
		There exist cost and transfer rules $\cC(\cdot , \cdot \, ; \nu_t)$ and $\cT(\cdot , \cdot \, ; \nu_t)$ such that, 
		if the GS algorithm is applied over $V(\cdot, \cdot \, ; \nu_t)$ at every $t \in [T]$,
		then stability, low regret, 
		fairness, and $\frac{1}{2}$-high social welfare are guaranteed for all $(\cA, \mu)$. 
	\end{theorem}
	Theorem \ref{thm:possibility_result} states that, with costs and/or transfers, it is possible to simultaneously guarantee all four desiderata.
	It implies that the impossibility result of Proposition \ref{prop:no_P_no_T} only holds when costs and transfers are not allowed. 
	We give intuition for the proof in Section \ref{sec:discussion}.
	Several questions arise, 
	such as, 
	what types of costs and transfers
	provide guarantees on stability and regret?
	Moreover, 
	do such costs and transfers arise in practice?
	In the following section, 
	we provide insights into these questions by studying a few examples.

\section{EXAMPLE SETTINGS}\label{sec:examples}

In the previous section, 
we found that, without costs and transfers, 
it is impossible to simultaneously guarantee stability and low optimal regret.
On the other hand, 
with costs and transfers,
it becomes possible to achieve all four desiderata described in Section \ref{sec:objectives}.
In this section,
we unpack several examples in order to understand when this outcome is possible as well as give intuition for when costs and transfers arise. 
As before, all proofs can be found in the Appendix.

We emphasize that our objective is not to issue recommendations on how to implement costs and transfers.
Rather, our goal is to highlight the role of costs and transfers in competitive settings.
Intuitively, transfers can be viewed as payments between agents (e.g., prices or distance traveled), 
and costs can be viewed as frictions (e.g., time expended or entry fees).
Neither are necessarily monetary, 
and both naturally surface in competitive settings. 
In the following examples, 
we study  three of many possible cost and transfer rules.

\subsection{Proportional cost with no transfer}\label{sec:prop_costs}

In this section, 
we study cost and transfer rules and show that, while appealing to the platform, they fail to guarantee that low regret holds alongside stability.
In the proportional-cost, no-transfer setting, 
\begin{align}
	\cC(a_1,a_2 ; \psi) &= \gamma \psi(a_1,a_2) 
	\qquad 
	\cT(a_1,a_2 ; \psi) = 0 ,   \label{eq:P_1} %
\end{align}
for all $a_1, a_2 \in \cA$
 and $\psi: \cA \times \cA^+ \rightarrow \bbR$, where $\gamma  \in [0,1]$. 
Intuitively, the amount that an agent $a_1$ expends is proportional to its desire to be matched to $a_2$. 
For example, the amount that a retailer pays a platform to have its advertisement shown to a user
or the amount of effort a worker puts into an internship
depends on the respective agent's level of interest. 
The following result shows that proportional costs do not improve upon the no-cost, no-transfer setting.

\begin{proposition} \label{prop:prop_costs}
	Suppose that $\cC$ and $\cT$ are set according to
	 \eqref{eq:P_1}.
	If the GS algorithm is applied over $V(\cdot, \cdot \, ; \nu_t)$ at every $t \in [T]$, then the system is stable. 
	If $\gamma \in [0,1)$, 
	$\uR (a ; \ccM) \leq 2 N^2 L (1 - \gamma)  \Delta_{\max} (a)   \left( \frac{8\sigma^2 \alpha \log(T)}{(1 - \gamma)^2 \Delta_{\min}^2} + \frac{\alpha}{\alpha - 2} \right)$
	and, if $\gamma = 1$, 
	$\uR(a ; \ccM) \leq 0$.
	However, under stability, it is not possible to guarantee fairness or high social welfare, and there exist settings $(\cA, \mu)$ for which $\bR(a ; \ccM) = \Omega(T)$ for at least one agent $a \in \cA$. 
\end{proposition}

Proposition \ref{prop:prop_costs} shows that this intuitively pleasing setting unfortunately yields the same negative result as  the no-cost, no-transfer setting. 
Interestingly, this setting is appealing to platforms, as shown next.

\begin{corollary} \label{cor:prop_cost_max_rew}
	Suppose that $\cC$ and $\cT$ are set according to \eqref{eq:P_1}
	and that the platform wishes to maximize  $F_t: \cW \rightarrow \bbR$ at every  $t \in [T]$. 
	Then, 
	when $\gamma = 1$, 
	there is no performance cost to stability: i.e., $\arg \max_{\sM \in \cW} F_t(\sM) \in S(V(\cdot, \cdot \, ; \nu_t))$ for all $t \in [T]$.
\end{corollary}

In other words, 
when $\gamma = 1$, 
the platform can maximize any objective function it desires while remaining 
confident that the users and providers will continue to use the platform's services. 
However, by the impossibility result given in Proposition \ref{prop:prop_costs}, the fact that users and providers continue to use the platform for matching does \emph{not} mean that they do so happily. 
Some agents may be unhappy because the platform's matchings
cause them to incur high optimal regret. 
Furthermore, they may incur high regret 
while other agents are allowed to incur low regret.
This finding suggests that such a platform may be barely holding onto its agents and may lose them if an alternate platform that offers slightly better outcomes arises. 
	
This setting also illustrates why it is  prudent to consider the agents' optimal regret in addition to their pessimal regret. 
By Proposition \ref{prop:prop_costs}, 
an agent's pessimal regret can be  non-positive while its optimal regret grows linearly in $T$. 
This finding reinforces the notion that it is not enough to upper bound the pessimal regret because there may be a very large gap between optimal and pessimal regrets.

\subsection{Balanced transfer}\label{sec:balanced_transfer}

We now introduce a transfer rule under which the platform can simultaneously guarantee all four desiderata.
In the no-cost and balanced-transfer setting, 
\begin{align}
	\cC(a_1,a_2 ; \psi) &= 0 ,  \label{eq:T_1} 
	\\
	\cT(a_1,a_2; \psi) &= 
	(\psi(a_2,a_1) - \psi(a_1,a_2)) / 2,   \label{eq:T_2} %
\end{align}
for all $a_1 , a_2 \in \cA$
 and $\psi: \cA \times \cA^+ \rightarrow \bbR$.  
To gain an intuition for this setting, suppose $\psi(a_1,a_2) > \psi(a_2, a_1)$ such that
agent $a_1$ benefits more from the match $(a_1,a_2)$ than $a_2$ does. 
Under \eqref{eq:T_1}, $a_1$ compensates for this imbalance by investing resources to make the experience more amenable for $a_2$. 
For example, suppose that an athlete $a_1$ is more interested in lessons with coach $a_2$ than $a_2$ is interested in coaching $a_1$. Then, $a_1$ can offset this difference by offering to pay more for lessons with $a_2$. 
In the literature, the balanced transfer is viewed as compensation or a bargaining solution.

The next result shows that the platform can simultaneously guarantee all four desiderata. 
Let 
\begin{align*}
	&\rho(a,a') = \frac{1}{2}( \mu(a,a') + \mu(a',a)) ,
	\\
	&\Delta^{\rho}_{\min} = \min_{ a_1 \in \cA,a_2 \in \cA^+, a_3 \in \cA^+ \setminus a_2} | \rho(a_1,a_2) -  \rho(a_1,a_3) | .
\end{align*}
\vspace{1pt}

\begin{theorem}\label{thm:balanced_transfer}
	Suppose that $\cC$ and $\cT$ are set according to \eqref{eq:T_1}-\eqref{eq:T_2}.
	Then, 
	$| S(\rho) | = 1$.
	Let $\sM^{*,\rho}$ be the only element in $S(\rho)$ 
	and
	$\Delta_{\max}^{*,\rho}(a) = \max_{a' \in \cA^+} (\rho(a,\sM^{*,\rho}(a)) - \rho(a,a') )$. 
	If the GS algorithm is applied over $V(\cdot, \cdot \, ; \nu_t)$ at every $t \in [T]$, then the system is stable, the social welfare is $\frac{1}{2}$-high at all $t \in [T]$,
		$\ccM$ is fair, and
	$\uR(a; \ccM) = \bR(a ; \ccM) \leq 
	\Delta_{\max}^{*,\rho}(a) N^2 L    \left( \frac{8 \sigma^2 \alpha \log(T)}{ (\Delta^{\rho}_{\min})^2} +  \frac{\alpha}{\alpha - 2} \right)$
	for all $a \in \cA$. 
\end{theorem}

Therefore,  under balanced transfers, applying the GS algorithm at every time step {simultaneously guarantees stability, low regret, fairness, and high social welfare}. 
Put differently, under balanced transfers between matched agents, 
if one guarantees that the matchings are stable, one gets low optimal regret, fairness, and high social welfare for free. 
One would not necessarily expect such a result to hold true when agents compete while learning (or, equivalently, learn while competing). 
For example, 
a college student hoping to learn their ideal career path through internships may be hurt 
by the presence of competition. 
It is not clear that an assignment could be stable and allow all agents to learn their preferences efficiently, let alone guarantee fairness and high social welfare.

\subsection{Pricing} \label{sec:MP}

In the previous section, we study a transfer rule in which the payment between agents depends on the preferences of both agents.
While such transfers are often feasible, 
there are settings in which the amount transferred cannot be tailored to the parties involved in the match. 
For example, suppose each provider is selling a product for which the price does not depend on which user purchases it.
To capture such contexts, we turn our attention to the pricing setting, in which: 
\begin{align}
	\cC(p,u ; \psi) &= c_1 
	\qquad  
	\cT(p,u ; \psi) = g(p ; \psi) 
	\qquad \label{eq:M_1}  
	\\
	\cC(u,p ; \psi) &= c_2 
	\qquad
	\cT(u,p ; \psi) = - g(p ; \psi) \label{eq:M_3}  
\end{align}
for all $u \in \cU$, $p \in \cP$, and $\psi : \cA \times \cA^+ \rightarrow \bbR$, where $c_1, c_2 \in \bbR$ and $g: \cP \rightarrow \bbR$.\footnote{
	Note that the presence of costs $c_1$ and $c_2$ is a technical detail. 
	For those interested, recall that, without loss of generality, we assume all agents prefer to be matched than unmatched, and $\cC(a, \emptyset) = \cT(a, \emptyset) = 0$ for all $a \in \cA$. 
	$c_1$ and $c_2$ are normalizing constants to ensure that this assumption is consistent with our analysis. 
	For instance, in one setting that we examine in the Appendix, $c_1 = 0$ and $c_2 < 0$.
	$-c_2$ can be viewed as a baseline utility that all agents gain if they are matched versus when they are unmatched. 
}
The transfer $g(p ; \psi)$ can be viewed as the ``price'' of being matched with $p$ that is constant across all agents wishing to ``purchase'' a match with $p$. 
This setting differs from previous ones in that $\cC$ and $\cT$ do not depend on the preferences of both matched agents.
Rather, based on $\nu_t$, $g(p ; \nu_t)$ is assigned to providers in a way that is reminiscent of assigning prices to goods.

For example, suppose $\cU$ is a set of patients, each hoping to book an appointment with a doctor in $\cP$. 
Suppose the price of an appointment with $p$ is fixed such that $g(p ; \nu_t)$ does not depend on the specific patient $u$. 
Since there are a limited number of available appointments, $g(p ; \nu_t)$ may depend on the current demand $\nu_t$, which can reflect, for instance, different patient needs and different doctor specialties. 
If an appointment is booked (i.e., a match is made), then the patient pays the doctor (i.e., makes a transfer).  
With time, patients learn their preferences over doctors, and doctors learn which patients require their services.
In such settings, fairness matters greatly, and preferences must be learned from a small number of sparse interactions.

We now show that, under mild conditions, there always exists at least one pricing rule that simultaneously guarantees stability, low regret, and fairness. 

\begin{theorem}\label{thm:pricing}
	Suppose that $\cC$ and $\cT$ are set according to \eqref{eq:M_1}-\eqref{eq:M_3}. 
	If the GS algorithm is applied over $V(\cdot, \cdot \, ; \nu_t)$ at every $t \in [T]$, then the system is stable. 
	Moreover, if  there exists a $B \in \bbR_{> 0}$ such that $| \mu(u, \cdot) | \leq B$ for all $u \in \cU$,
	then 
	$| S(V(\cdot, \cdot \, ; \mu)) | = 1$
	and
	there exist constants $c_1$ and $c_2$ as well as a pricing rule $g$ such that
	$\ccM$ is fair
	and 
	$\uR (a; \ccM) = \bR(a ; \ccM) \leq 
	2 \Delta_{\max}^{*,{\scriptscriptstyle B}}(a) N^2 L \left( \frac{8 \sigma^2 \alpha \log(T)}{ (\Delta_{\min} )^2} +  \frac{\alpha}{\alpha - 2} \right)$
	for all $a \in \cA$, 
	where
	$\sM^{*,{\scriptscriptstyle B}}$ is the only element in $S(V(\cdot, \cdot \, ; \mu))$
	and $\Delta^{*,{\scriptscriptstyle B}}_{\max}(a) = 2 B(L - 1) \mathbf{1}(a \in \cU) + \max_{a' \in \cA^+} ( \mu(a, \sM^{*,{\scriptscriptstyle B}}(a) )  - \mu(a, a' ))$.
\end{theorem}
This result shows that, under mild conditions, there is always at least one way to price matches such that the platform {simultaneously guarantees stability, low optimal regret, and fairness}. 
Put differently, one can always find a pricing rule such that, if the matching is stable, low optimal regret and fairness come for free. 
Under the pricing rule with which we show existence in the proof, high social welfare is not guaranteed, but it may be possible under other pricing rules. 

Finally, 
note that the results in this section hold with small modifications, such as if users and providers are switched; if $\cC(u,p ; \psi) = g(p ; \psi) - c_1$, $\cC(p,u ; \psi) = 0$, and $\cT = 0$; or if $\cC(u,p ; \psi) = c_2  + (1 - \gamma') g(p ; \psi)$, $\cT(u,p ; \psi) = - \gamma ' g(p ; \psi)$, $\cC(p,u ; \psi) = c_1 - (1 - \gamma') g(p ; \psi)$, and $cT(p,u ; \psi) = \gamma' g(p ; \psi)$.

	\section{DISCUSSION} \label{sec:discussion}
	\vspace{-2pt}
	
	\noindent \textbf{Summary}.
	In this work, we model matching under learning by blending the matching and MAB problems. 
	As in the two-sided matching problem, users compete for matches with providers, and vice versa. 
	As in the MAB problem, agents do not know their preferences \emph{a priori} and must learn them from experience. 
	
	We focus our attention on the \emph{centralized matching market with bandit learners}, extending the results of \citet{liu2020competing}.
	In particular, the framework offered by \citet{liu2020competing} suggests that it is not possible for competitive matching markets to simultaneously guarantee stability and low optimal regret (a measure of long-term happiness). 
	This impossibility result could have harmful consequences, 
	such as providing a justification for platforms to treat some of its agents well while treating others poorly.

	In this work, we show that this impossibility does not always hold. 
	We show that, if we account for two natural components of competition---costs and transfers---it
	is possible to {simultaneously guarantee four desiderata}: 
	(1) \emph{stability} such that no pair of agents is incentivized to defect from the matching; (2) \emph{low regret} such that all agents incur $O(\log(T))$ optimal and pessimal regret; (3) the \emph{fair} distribution of regret across agents; and (4) high \emph{social welfare}.

	\noindent \textbf{Proof intuition}. 
	Ensuring $(\ccM, \cC, \cT)$ satisfies the first three desiderata consists of three main ingredients. 
	First, the GS algorithm is applied over $V(\cdot, \cdot \, ; \nu_t)$ at every time step $t \in [T]$ to guarantee stability. 
	Second, $\cC$ and $\cT$ are chosen such that, under the appropriate conditions, $|S(V(\cdot, \cdot \, ; \psi)) | = 1$ for any $\psi$. 
	In other words, $\cC$ and $\cT$ yield a \emph{unique} stable matching given preferences $\psi$. 
	Recalling the definition of regret in Section \ref{sec:objectives}, 
	uniqueness ensures that 
	it is not possible for the following to occur: the system converges to a stable matching $\sM \in S(V(\cdot, \cdot \, ; \mu))$ for which one agent incurs $O(\log(T))$ regret at the expense of another agent $a$ 
	incurring $\Omega(T)$ regret 
	because $a$'s  optimal stable matching is $\sM' \in S(V(\cdot, \cdot \, ; \mu)) \setminus \sM$. 
	Finally,
	one must be careful that the rules $\cC$ and $\cT$ that induce a unique stable matching do not interfere with agents'
	abilities to learn their preferences $\mu$, 
	as discussed next.

	\textbf{Uniqueness is not enough}. 
	Selecting $(\mathcal{M}, \mathcal{C}, \mathcal{T})$ such that  there is a unique stable matching implies that the user- and provider-optimal stable matchings are identical. 
	It therefore implies that the optimal and pessimal regret are the same. 
	What it does \emph{not} imply, is that the optimal and pessimal regret are $O(\log(T))$. 
	One may be tempted to draw the following conclusion:
	Since Liu et al. (2020) show that pessimal regret is $O(\log(T))$, 
	the optimal regret should also be $O(\log(T))$ when there is a unique stable matching. 
	
	However,
	drawing this conclusion would be a mistake. 
	Why? 
	Because the very tool used to induce a unique stable matching---specifically, $\cC$ and $\cT$---can interfere  efficient learning and prevent the system from converging to a matching that is stable under $\mu$. 
	In other words, when there are costs and transfers, 
	even the pessimal regret is not guaranteed to be $O(\log(T))$. 
	The key observation here is that the $O(\log(T))$ bound on pessimal regret given by Liu et al. (2020)---and re-stated in Proposition \ref{prop:no_P_no_T}---is proven for a setting in which $\cC$ and $\cT$ do not exist.
	When $\mathcal{C}$ and $\mathcal{T}$ are used to induce a unique stable matching, 
	it takes extra care to ensure that $\mathcal{C}$ and $\mathcal{T}$ do not interfere with learning in the process. 
	Furthermore, 
	recall that $\mu$ is unknown and only $\nu_t$ is available.
	Although the goal is to arrive at a matching that is stable under $\mu$ in $O(\log(T))$ time steps, 
	one must do so without knowledge of $\mu$. 
	Instead, $\cC$ and $\cT$ must be enforced with respect to the time-varying preferences $\nu_t$. 
	
	\noindent \textbf{Takeaways for matching platforms}. 
	This work shows that, under costs and transfers, one can simultaneously guarantee four desiderata of interest. 
	We provide one such rule---the balanced transfer---in Section \ref{sec:balanced_transfer}. 
	We provide another example in Section \ref{sec:MP} in which the costs and transfers are reminiscent of the pricing of goods. 
	However, there are many more cost and transfer rules that may be of interest. 
	
	One of the takeaways of this work is that it is possible to guarantee stability \emph{without} ensuring low  regret, as shown in Proposition \ref{prop:no_P_no_T} and Section  \ref{sec:prop_costs}. 
	This fact indicates that the
	continued participation of agents (e.g., social media users) is \emph{not} a sign that agents are satisfied with the platform's matchings. 
	Indeed, as discussed in Section \ref{sec:prop_costs}, agents can incur high regret even if they do not leave the platform
	because, by definition, stability implies that agents are not incentivized to defect. 
	It may be to the long-term benefit of a platform to remedy this issue since it leaves room for other platforms that guarantee low regret alongside stability to attract agents away from the platform. 
	
	\noindent \textbf{Future work}. 
	There are many possible paths for future work,
	such as considering strategy-proofness 
	or 
	allowing information sharing. 
	One could also study
	related problems with bandit learners under costs and transfers, 
	including decentralized matchings, 
	matchings with ties, 
	the hospital-residents problem (in which multiple users can be matched to the same provider), and more.
	In addition to the costs and transfers examined in Sections \ref{sec:examples},
	 one could characterize the full set of cost and transfer rules under which each desideratum is achievable. 
	Another research direction would be to apply and test the theoretical findings of this work on real-world systems,
	such as online labor markets or recommender systems.

\newpage

\subsubsection*{Acknowledgements}

We would like to thank our reviewers for their time and thoughtful comments. 
We would also like to thank Meena Jagadeesan, Alexander Wei, Karthik A. Sastry, Zachary Schiffer, and especially Horia Mania for their feedback. 
This work was supported in parts by the MIT-IBM project on ``Representation
Learning as a Tool for Causal Discovery'', the NSF TRIPODS Phase II grant towards Foundations of
Data Science Institute, the Chyn Duog Shiah Memorial Fellowship, and the Hugh Hampton Young
Memorial Fund Fellowship.

\bibliography{ref.bib}

\begin{thebibliography}{51}
\providecommand{\natexlab}[1]{#1}
\providecommand{\url}[1]{\texttt{#1}}
\expandafter\ifx\csname urlstyle\endcsname\relax
  \providecommand{\doi}[1]{doi: #1}\else
  \providecommand{\doi}{doi: \begingroup \urlstyle{rm}\Url}\fi

\bibitem[Aridor et~al.(2020)Aridor, Mansour, Slivkins, and
  Wu]{aridor2020competing}
Guy Aridor, Yishay Mansour, Aleksandrs Slivkins, and Zhiwei~Steven Wu.
\newblock Competing {B}andits: The {P}erils of {E}xploration {U}nder
  {C}ompetition.
\newblock \emph{arXiv:2007.10144 [cs.GT]}, 2020.

\bibitem[Ashlagi and Gonczarowski(2018)]{ashlagi2018stable}
Itai Ashlagi and Yannai~A. Gonczarowski.
\newblock Stable matching mechanisms are not obviously strategy-proof.
\newblock \emph{Journal of Economic Theory}, 177:\penalty0 405--425, 2018.

\bibitem[Auer et~al.(2002)Auer, Cesa-Bianchi, and Fischer]{auer2002finite}
Peter Auer, Nicol\`{o} Cesa-Bianchi, and Paul Fischer.
\newblock Finite-time {A}nalysis of the {M}ultiarmed {B}andit {P}roblem.
\newblock \emph{Machine learning}, 47\penalty0 (2):\penalty0 235--256, 2002.

\bibitem[Avner and Mannor(2019)]{avner2019multi}
Orly Avner and Shie Mannor.
\newblock Multi-{U}ser {C}ommunication {N}etworks: A {C}oordinated
  {M}ulti-{A}rmed {B}andit {A}pproach.
\newblock \emph{IEEE/ACM Transactions on Networking}, 27\penalty0 (6):\penalty0
  2192--2207, 2019.

\bibitem[Axtell and Kimbrough(2008)]{axtell2008high}
Robert~L. Axtell and Steven~O. Kimbrough.
\newblock The {H}igh {C}ost of {S}tability in {T}wo-{S}ided {M}atching: How
  {M}uch {S}ocial {W}elfare {S}hould be {S}acrificed in the {P}ursuit of
  {S}tability.
\newblock In \emph{Proceedings of the 2008 World Congress on Social Simulation
  (WCSS 2008)}, Fairfax, USA, 2008.

\bibitem[Becker(1973)]{becker1973theory}
Gary~S. Becker.
\newblock A {T}heory of {M}arriage: Part {I}.
\newblock \emph{Journal of Political Economy}, 81\penalty0 (4):\penalty0
  813--846, 1973.

\bibitem[Boursier and Perchet(2020)]{boursier2020selfish}
Etienne Boursier and Vianney Perchet.
\newblock Selfish {R}obustness and {E}quilibria in {M}ulti-{P}layer {B}andits.
\newblock In Jacob Abernethy and Shivani Agarwal, editors, \emph{Proceedings of
  Thirty Third Annual Conference on Learning Theory (COLT 2020)}, pages
  530--581, Online, 2020. PMLR.

\bibitem[Bubeck and Cesa-Bianchi(2012)]{bubeck2012regret}
S{\'e}bastien Bubeck and Nicol\`{o} Cesa-Bianchi.
\newblock Regret {A}nalysis of {S}tochastic and {N}onstochastic {M}ulti-armed
  {B}andit {P}roblems.
\newblock \emph{Foundations and Trends in Machine Learning}, 5\penalty0 (1),
  2012.

\bibitem[Bubeck et~al.(2020)Bubeck, Li, Peres, and Sellke]{bubeck2020non}
S{\'e}bastien Bubeck, Yuanzhi Li, Yuval Peres, and Mark Sellke.
\newblock Non-{S}tochastic {M}ulti-{P}layer {M}ulti-{A}rmed {B}andits: Optimal
  {R}ate {W}ith {C}ollision {I}nformation, {S}ublinear {W}ithout.
\newblock In Jacob Abernethy and Shivani Agarwal, editors, \emph{Proceedings of
  Thirty Third Annual Conference on Learning Theory (COLT 2020)}, volume 125,
  pages 961--987, Online, 2020. PMLR.

\bibitem[Cesa-Bianchi and Lugosi(2006)]{cesa2006prediction}
Nicol{`o} Cesa-Bianchi and G{\'a}bor Lugosi.
\newblock \emph{Prediction, {L}earning, and {G}ames}.
\newblock Cambridge University Press, New York, NY, USA, 2006.

\bibitem[Chawla et~al.(2020)Chawla, Sankararaman, Ganesh, and
  Shakkottai]{chawla2020gossiping}
Ronshee Chawla, Abishek Sankararaman, Ayalvadi Ganesh, and Sanjay Shakkottai.
\newblock The {G}ossiping {I}nsert-{E}liminate {A}lgorithm for {M}ulti-{A}gent
  {B}andits.
\newblock In Silvia Chiappa and Roberto Calandra, editors, \emph{Proceedings of
  the Twenty Third International Conference on Artificial Intelligence and
  Statistics (AISTATS 2020)}, pages 3471--3481, Online, 2020. PMLR.

\bibitem[Das and Kamenica(2005)]{das2005two}
Sanmay Das and Emir Kamenica.
\newblock Two-{S}ided {B}andits and the {D}ating {M}arket.
\newblock In Leslie~Pack Kaelbling, editor, \emph{Proceedings of the Nineteenth
  International Joint Conference on Artificial Intelligence (IJCAI 2005)},
  volume~5, pages 947--952, Edinburgh, Scotland, 2005.

\bibitem[Dean and Swar(2009)]{dean2009generalized}
Brian~C Dean and Namrata Swar.
\newblock The {G}eneralized {S}table {A}llocation {P}roblem.
\newblock In \emph{Proceedings of Third International Workshop on Algorithms
  and Computation (WALCOM 2009)}, volume 5431, pages 238--249, Kolkata, India,
  2009. Springer-Verlag.

\bibitem[Echenique and Yariv(2012)]{echenique2012experimental}
Federico Echenique and Leeat Yariv.
\newblock An {E}xperimental {S}tudy of {D}ecentralized {M}atching.
\newblock Technical report, California Institute of Technology, Pasadena, CA,
  USA, 2012.

\bibitem[Gale and Shapley(1962)]{gale1962college}
David Gale and Lloyd~S. Shapley.
\newblock College {A}dmissions and the {S}tability of {M}arriage.
\newblock \emph{The American Mathematical Monthly}, 69\penalty0 (1):\penalty0
  9--15, 1962.

\bibitem[Hitsch et~al.(2010)Hitsch, Horta{\c{c}}su, and
  Ariely]{hitsch2010matching}
Gunter~J. Hitsch, Ali Horta{\c{c}}su, and Dan Ariely.
\newblock Matching and {S}orting in {O}nline {D}ating.
\newblock \emph{The American Economic Review}, 100\penalty0 (1):\penalty0
  130--63, 2010.

\bibitem[Irving(1985)]{irving1985efficient}
Robert~W. Irving.
\newblock An {E}fficient {A}lgorithm for the “{S}table {R}oommates”
  {P}roblem.
\newblock \emph{Journal of Algorithms}, 6\penalty0 (4):\penalty0 577--595,
  1985.

\bibitem[Irving(1994)]{irving1994stable}
Robert~W. Irving.
\newblock {S}table marriage and indifference.
\newblock \emph{Discrete Applied Mathematics}, 48\penalty0 (3):\penalty0
  261--272, 1994.

\bibitem[Jagadeesan et~al.(2021)Jagadeesan, Wei, Wang, Jordan, and
  Steinhardt]{jagadeesan2021learning}
Meena Jagadeesan, Alexander Wei, Yixin Wang, Michael~I Jordan, and Jacob
  Steinhardt.
\newblock Learning equilibria in matching markets from bandit feedback.
\newblock \emph{arXiv preprint arXiv:2108.08843}, 2021.

\bibitem[Kalathil et~al.(2014)Kalathil, Nayyar, and
  Jain]{kalathil2014decentralized}
Dileep Kalathil, Naumaan Nayyar, and Rahul Jain.
\newblock Decentralized {L}earning for {M}ultiplayer {M}ultiarmed {B}andits.
\newblock \emph{IEEE Transactions on Information Theory}, 60\penalty0
  (4):\penalty0 2331--2345, 2014.

\bibitem[Kelso~Jr. and Crawford(1982)]{kelso1982job}
Alexander~S. Kelso~Jr. and Vincent~P. Crawford.
\newblock Job {M}atching, {C}oalition {F}ormation, and {G}ross {S}ubstitutes.
\newblock \emph{Econometrica}, 50\penalty0 (6):\penalty0 1483--1504, 1982.

\bibitem[Klaus and Klijn(2006)]{klaus2006procedurally}
Bettina Klaus and Flip Klijn.
\newblock Procedurally fair and stable matching.
\newblock \emph{Economic Theory}, 27\penalty0 (2):\penalty0 431--447, 2006.

\bibitem[Knuth and De~Bruijn(1997)]{knuth1997stable}
Donald~Ervin Knuth and N.~G. De~Bruijn.
\newblock \emph{Stable {M}arriage and {I}ts {R}elation to {O}ther
  {C}ombinatorial {P}roblems: An {I}ntroduction to the {M}athematical
  {A}nalysis of {A}lgorithms}, volume~10 of \emph{CRM proceedings \& lecture
  notes}.
\newblock American Mathematical Society, 1997.

\bibitem[Kuhn(1955)]{kuhn1955hungarian}
Harold~W. Kuhn.
\newblock The {H}ungarian method for the assignment problem.
\newblock \emph{Naval Research Logistics Quarterly}, 2\penalty0 (1-2):\penalty0
  83--97, 1955.

\bibitem[Lai and Robbins(1985)]{lai1985asymptotically}
T.~L. Lai and Herbert Robbins.
\newblock Asymptotically efficient adaptive allocation rules.
\newblock \emph{Advances in Applied Mathematics}, 6\penalty0 (1):\penalty0
  4--22, 1985.

\bibitem[Lattimore and Szepesv{\'a}ri(2020)]{lattimore2020bandit}
Tor Lattimore and Csaba Szepesv{\'a}ri.
\newblock \emph{Bandit {A}lgorithms}.
\newblock Cambridge University Press, Cambridge, UK, 2020.

\bibitem[Liu and Zhao(2010)]{liu2010distributed}
Keqin Liu and Qing Zhao.
\newblock Distributed {L}earning in {M}ulti-{A}rmed {B}andit {W}ith {M}ultiple
  {P}layers.
\newblock \emph{IEEE Transactions on Signal Processing}, 58\penalty0
  (11):\penalty0 5667--5681, 2010.

\bibitem[Liu et~al.(2020{\natexlab{a}})Liu, Mania, and
  Jordan]{liu2020competing}
Lydia~T. Liu, Horia Mania, and Michael~I. Jordan.
\newblock Competing {B}andits in {M}atching {M}arkets.
\newblock In Silvia Chiappa and Roberto Calandra, editors, \emph{Proceedings of
  the Twenty Third International Conference on Artificial Intelligence and
  Statistics (AISTATS 2020)}, pages 1618--1628, Online, 2020{\natexlab{a}}.
  PMLR.

\bibitem[Liu et~al.(2020{\natexlab{b}})Liu, Ruan, Mania, and
  Jordan]{liu2020bandit}
Lydia~T. Liu, Feng Ruan, Horia Mania, and Michael~I. Jordan.
\newblock Bandit {L}earning in {D}ecentralized {M}atching {}markets.
\newblock \emph{arXiv:2012.07348 [cs.LG]}, 2020{\natexlab{b}}.

\bibitem[Manlove(2002)]{manlove2002structure}
David~F. Manlove.
\newblock The structure of stable marriage with indifference.
\newblock \emph{Discrete Applied Mathematics}, 122\penalty0 (1-3):\penalty0
  167--181, 2002.

\bibitem[Mansour et~al.(2018)Mansour, Slivkins, and Wu]{mansour2018competing}
Yishay Mansour, Aleksandrs Slivkins, and Zhiwei~Steven Wu.
\newblock Competing {B}andits: {L}earning {U}nder {C}ompetition.
\newblock In Anna~R. Karlin, editor, \emph{9th Innovations in Theoretical
  Computer Science Conference (ITCS 2018)}, volume~94 of \emph{Leibniz
  International Proceedings in Informatics (LIPIcs)}, pages 48:1--48:27,
  Dagstuhl, Germany, 2018. Schloss Dagstuhl--Leibniz-Zentrum fuer Informatik.

\bibitem[Masarani and Gokturk(1989)]{masarani1989existence}
F.~Masarani and Sadik~S. Gokturk.
\newblock On the existence of fair matching algorithms.
\newblock \emph{Theory and Decision}, 26\penalty0 (3):\penalty0 305--322, 1989.

\bibitem[Pentico(2007)]{pentico2007assignment}
David~W. Pentico.
\newblock Assignment problems: A golden anniversary survey.
\newblock \emph{European Journal of Operational Research}, 176\penalty0
  (2):\penalty0 774--793, 2007.

\bibitem[Ross and Soland(1975)]{ross1975branch}
G.~Terry Ross and Richard~M. Soland.
\newblock A branch and bound algorithm for the generalized assignment problem.
\newblock \emph{Mathematical Programming}, 8:\penalty0 91--103, 1975.

\bibitem[Roth(1982)]{roth1982economics}
Alvin~E. Roth.
\newblock The {E}conomics of {M}atching: {S}tability and {I}ncentives.
\newblock \emph{Mathematics of Operations Research}, 7\penalty0 (4):\penalty0
  617--628, 1982.

\bibitem[Roth(1984)]{roth1984evolution}
Alvin~E. Roth.
\newblock The {E}volution of the {L}abor {M}arket for {M}edical {I}nterns and
  {R}esidents: A {C}ase {S}tudy in {G}ame {T}heory.
\newblock 92\penalty0 (6):\penalty0 991--1016, 1984.

\bibitem[Roth(2008)]{roth2008deferred}
Alvin~E Roth.
\newblock Deferred acceptance algorithms: {h}istory, theory, practice, and open
  questions.
\newblock \emph{International Journal of Game Theory}, 36:\penalty0 537--569,
  2008.

\bibitem[Roth and Vate(1990)]{roth1990random}
Alvin~E. Roth and John H.~Vande Vate.
\newblock Random {P}aths to {S}tability in {T}wo-{S}ided {M}atching.
\newblock \emph{Econometrica}, 58\penalty0 (6):\penalty0 1475--1480, 1990.

\bibitem[Sankararaman et~al.(2020)Sankararaman, Basu, and
  Sankararaman]{sankararaman2020dominate}
Abishek Sankararaman, Soumya Basu, and Karthik~Abinav Sankararaman.
\newblock Dominate or {D}elete: {D}ecentralized {C}ompeting {B}andits with
  {U}niform {V}aluation.
\newblock \emph{arXiv:2006.15166 [cs.LG]}, 2020.

\bibitem[Schnabel et~al.(2018)Schnabel, Bennett, Dumais, and
  Joachims]{schnabel2018short}
Tobias Schnabel, Paul~N. Bennett, Susan~T. Dumais, and Thorsten Joachims.
\newblock Short-{T}erm {S}atisfaction and {L}ong-{T}erm {C}overage:
  Understanding {H}ow {U}sers {T}olerate {A}lgorithmic {E}xploration.
\newblock In \emph{Proceedings of the Eleventh ACM International Conference on
  Web Search and Data Mining (WSDM 2018)}, pages 513--521, Marina Del Rey, CA,
  USA, 2018.

\bibitem[Shapley and Scarf(1974)]{shapley1974cores}
Lloyd Shapley and Herbert Scarf.
\newblock On cores and indivisibility.
\newblock \emph{Journal of Mathematical Economics}, 1:\penalty0 23--37, 1974.

\bibitem[Shapley and Shubik(1971)]{shapley1971assignment}
Lloyd~S. Shapley and Martin Shubik.
\newblock The {A}ssignment {G}ame {I}: The {C}ore.
\newblock \emph{International Journal of Game theory}, 1\penalty0 (1):\penalty0
  111--130, 1971.

\bibitem[S{\"u}hr et~al.(2019)S{\"u}hr, Biega, Zehlike, Gummadi, and
  Chakraborty]{suhr2019two}
Tom S{\"u}hr, Asia~J. Biega, Meike Zehlike, Krishna~P. Gummadi, and Abhijnan
  Chakraborty.
\newblock Two-{S}ided {F}airness for {R}epeated {M}atchings in {T}wo-{S}ided
  {M}arkets: A {C}ase {S}tudy of a {R}ide-{H}ailing {P}latform.
\newblock In \emph{Proceedings of the Twenty Fifth ACM SIGKDD International
  Conference on Knowledge Discovery \& Data Mining}, pages 3082--3092, New
  York, NY, USA, 2019. Association for Computing Machinery.

\bibitem[Sutton and Barto(2018)]{sutton2018reinforcement}
Richard~S. Sutton and Andrew~G. Barto.
\newblock \emph{Reinforcement {L}earning: An {I}ntroduction}.
\newblock The MIT Press, 2nd edition, 2018.

\bibitem[Thompson(1933)]{thompson1933likelihood}
William~R. Thompson.
\newblock On the {L}ikelihood that {O}ne {U}nknown {P}robability {E}xceeds
  {A}nother in {V}iew of the {E}vidence of {T}wo {S}amples.
\newblock \emph{Biometrika}, 25\penalty0 (3-4):\penalty0 285--294, 1933.

\bibitem[Vershynin(2018)]{vershynin2018high}
Roman Vershynin.
\newblock \emph{High-{D}imensional {P}robability: An {I}ntroduction with
  {A}pplications in {D}ata {S}cience}, volume~47 of \emph{Cambridge Series in
  Statistical and Probabilistic Mathematics}.
\newblock Cambridge University Press, Cambridge, UK, 2018.

\bibitem[Vial et~al.(2020)Vial, Shakkottai, and Srikant]{vial2020robust}
Daniel Vial, Sanjay Shakkottai, and R.~Srikant.
\newblock Robust {M}ulti-{A}gent {M}ulti-{A}rmed {B}andits.
\newblock \emph{arXiv:2007.03812 [cs.LG]}, 2020.

\bibitem[Wiering and Van~Otterlo(2012)]{wiering2012reinforcement}
Marco Wiering and Martijn Van~Otterlo.
\newblock Reinforcement {L}earning: State-of-the-{A}rt.
\newblock \emph{Adaptation, Learning, and Optimization}, 12\penalty0 (3), 2012.

\bibitem[Wilson et~al.(2014)Wilson, Geana, White, and Ludvig]{wilson2014humans}
Robert~C. Wilson, Andra Geana, John~M. White, and Jonathan~D. Ludvig, Elliot A
  .and~Cohen.
\newblock Humans use directed and random exploration to solve the
  explore--exploit dilemma.
\newblock \emph{Journal of Experimental Psychology: General}, 143\penalty0
  (6):\penalty0 2074--2081, 2014.

\bibitem[Wilson et~al.(2021)Wilson, Bonawitz, Costa, and
  Ebitz]{wilson2020balancing}
Robert~C. Wilson, Elizabeth Bonawitz, Vincent~D. Costa, and R.~Becket Ebitz.
\newblock Balancing exploration and exploitation with information and
  randomization.
\newblock \emph{Current Opinion in Behavioral Sciences}, 38:\penalty0 49--56,
  2021.

\bibitem[Wu and Roth(2018)]{wu2018lattice}
Qingyun Wu and Alvin~E. Roth.
\newblock The lattice of envy-free matchings.
\newblock \emph{Games and Economic Behavior}, 109:\penalty0 201--211, 2018.

\end{thebibliography}

\clearpage
\appendix

\thispagestyle{empty}

\onecolumn \makesupplementtitle

\setcounter{theorem}{0}

All results that appear in the manuscript are numbered in the same manner below. 
Results that appear in the Appendix but not in the manuscript are preceded by the letter of the corresponding section.

\section{TECHNICAL DETAILS}

We re-state the GS algorithm given in Algorithm \ref{alg:GS_short} in detail below. 
In this algorithm, the providers are the proposers. 
The version of the algorithm in which users are the proposers is given in Appendix \ref{sec:app_balanced_transfer}.

{\centering
	\begin{minipage}{.95\linewidth}
\begin{algorithm}[H]
	\SetAlgoLined
	\KwIn{Set of agents $\cA = \cU \cup \cP$, where the set of users is  $\cU = \{u_1, u_2, \hdots, u_N \}$, the set of providers is $\cP  = \{ p_1, p_2, \hdots, p_L \}$, $\cU \cap \cP = \emptyset$, and $N \geq L$. A payoff function $V: \cA \times \cA^+ \rightarrow \bbR$.
	}
	\KwOut{Matching $\sM \in S(V) \subset \cW$. }
	\BlankLine
	Initialize matching $\sM: \cA \rightarrow \cA^+$ such that $\sM(a) \leftarrow \emptyset$ for all $a \in \cA$\;
	Initialize empty (FIFO) queues $Q(p) \leftarrow [ \hspace{1pt} ]$ for all $p \in \cP$\;
	\BlankLine
	\tcp{Fill each provider's queue with users in order of decreasing preference.}
	\For{$p \in \cP$}{
		\For{$i = 1, 2, \hdots, N$}{
			Append $r^{-1}(i ; V(p, \cdot))$ to $Q(p)$\tcp*[l]{Add $p$'s $i$-th ranked user.}
		}
	}
	
	\BlankLine
	\tcp{As long as there exists a provider who is unmatched...}
	\While{$\exists p \in \cP : \sM(p) = \emptyset$}{ \label{line:pick_proposer}
		$u \leftarrow \text{pop}(Q(p))$\tcp*[l]{Provider $p$'s favorite user of those remaining in $p$'s queue.}
		\BlankLine
		\tcp{If user $u$ is unmatched, match $u$ and $p$.}
		\uIf{$\sM(u) = \emptyset$}{
			$\sM(u) \leftarrow p$\;
			$\sM(p) \leftarrow u$\;
		}
		\BlankLine
		\tcp{If user $u$ prefers $p$ to its current match $\sM(u)$, match $u$ and $p$.}
		\uElseIf{$V(u,p ) > V(u , \sM(u) )$}{
			$\sM'( \sM(u) ) \leftarrow \emptyset$\;
			$\sM(u) \leftarrow p$\;
			$\sM(p) \leftarrow u$\;
		}
	}
	Return $\sM$\;
	\caption{Gale-Shapley algorithm (with providers as proposers)} \label{alg:GS}
	\algorithmfootnote{In the GS algorithm presented here, the providers are the proposers. The version of the algorithm in which users are the proposers is given in Appendix \ref{sec:app_balanced_transfer}.} 
\end{algorithm}
	\end{minipage}
\par
}

\textbf{Shorthands}. 
In this Appendix, we make use of the following notational shorthands. 
Let $\psi :\cA \times \cA^+ \rightarrow  \bbR$ denote an arbitrary function (e.g., preferences). 
We say that $a_1 \succ_{\psi(a_3, \cdot)} a_2$ if $\psi(a_3, a_1) > \psi(a_3, a_2)$.
We denote the rank---or ordinal preference---of agent $a'$ under $\psi(a, \cdot)$ by $r(a'; \psi(a, \cdot) )$, where $r : \cA^+ \rightarrow \bbN_{> 0}$ 
and $r(a'; \psi(a, \cdot)) < r( a''; \psi(a, \cdot)) \iff \psi(a, a') > \psi(a, a'')$.
For example, if $p = \argmax_{p \in \cP} \psi(u,p)$, then $r(p ; \psi(u, \cdot) ) = 1$. 
Accordingly, $r^{-1}(n ; \psi(a, \cdot))$ is the $n$-th highest ranked user under $\psi(a, \cdot)$.

\newpage

	\noindent \textbf{Assumptions}. At various points in our analysis, we make  uniqueness assumptions.
These assumptions are fairly mild, as they require that  the corresponding real-valued preferences (or payoffs) differ by some amount, however small. 
Intuitively, an assumption of uniqueness is equivalent to an assumption that an agent has a strict ordering of preferences (or payoffs).
This assumption is therefore not valid when there are ties (i.e., indifference) in an agent's preferences (or payoffs).
We can also relax
one of our assumptions that 
$\mu(a,a') + \mu(a',a) \neq \mu(a''',a'') + \mu(a'',a''')$ unless $(a, a') = (a'', a''')$ or $(a, a') = (a'', a''')$.
This assumption is only relevant to Theorem \ref{thm:possibility_result} and Theorem \ref{thm:balanced_transfer}.
Even so, it is not strictly necessary.
The assumption could be removed by introducing a consistent tie-breaking rule that the platform uses to choose between matchings, as discussed in Section \ref{sec:app_balanced_transfer}.
Because this setup requires additional notation, we opt for the more straightforward construction above.

\section{PROOFS FOR STABILITY RESULTS} \label{sec:app_GS}
\setcounter{theorem}{0}

\begin{lemmaApp}\label{lem:GS}
	Let $\psi: \cA \times \cA^+ \rightarrow \bbR$ denote an arbitrary preference function. 
	Let $\ccM(\cdot \, ; \psi)$ denote a matching returned by the GS algorithm, as given in Algorithm \ref{alg:GS}, when applied over payoffs $V(\cdot, \cdot \, ; \psi)$.
	Then, $\ccM(\cdot \, ; \psi)$ is a stable matching under payoffs $V(\cdot, \cdot \, ; \psi)$: i.e., $\ccM(\cdot \, ; \psi) \in S(V(\cdot, \cdot \, ; \psi))$. 
\end{lemmaApp}
\begin{proof}
	This is a well-known result, cf. \cite{gale1962college} or \cite{roth2008deferred}.
	To prove it, consider Algorithm \ref{alg:GS}. 
	Suppose that the result is not true, in which case there exists a $p_1 \in \cP$, $p_2 \in \cP \cup \emptyset$, and $u_1, u_2 \in \cU$ such that $\ccM(u_1; \psi ) = p_1$ and $\ccM(u_2 ; \psi) = p_2$, but 
	$u_2 \succ_{V(p_1 , \cdot \, ; \psi)} u_1$ and $p_1 \succ_{V(u_2 , \cdot \, ; \psi)} p_2$ by the definition of stability (see Section \ref{sec:objectives}).
	
	$u_2 \succ_{V(p_1 , \cdot \, ; \psi)} u_1$ implies that $p_1$ proposes to $u_2$ before proposing to $u_1$. 
	However, since $p_1 \succ_{V(u_2 , \cdot \, ; \psi)} p_2$, it is not possible that (a) $u_2$ rejects $p_1$ in favor of $p_2$ or (b) $u_2$ rejects $p_1$ in favor of $p_3 \in \cP \setminus \{p_1, p_2\}$ but that $u_2$ does not reject $p_2$ in favor of $p_3$. 
	Therefore, by contradiction, the GS algorithm returns a stable matching. 
\end{proof}

\begin{lemmaApp}\label{lem:GS_defection_proof}
	Consider the setup in Section \ref{sec:setup}. 
	Suppose that, at every time step $t \in [T]$, $\ccM( \cdot \, ; \nu_t)$ is a matching returned by the GS algorithm using the payoffs $V(\cdot, \cdot \, ; \nu_t)$. 
	Then, $(\ccM, \cC, \cT)$ is stable. 
\end{lemmaApp}
\begin{proof}
	This result follows from Lemma \ref{lem:GS} and Definition \ref{def:defection_proof}.
\end{proof}

In the remaining analysis, we will also make use of the following definition and lemma. 

\begin{definitionApp}\label{def:pessimal_optimal}
	A stable matching $\sM \in S(V)$ is \emph{$\cG$-optimal} under payoffs $V: \cA \times \cA^+ \rightarrow \bbR$ if $V(a, \sM(a)) \geq V(a, \sM'(a))$ for all $\sM'  \in S(V)$ and $a \in \cG$. 
	A stable matching $\sM \in S(V)$ is \emph{$\cG$-pessimal} under payoffs $V: \cA \times \cA^+ \rightarrow \bbR$ if $V(a, \sM(a)) \leq V(a, \sM'(a))$ for all $\sM'  \in S(V)$ and $a \in \cG$. 
\end{definitionApp}

\begin{lemmaApp} \label{lem:GS_proposer_optimal}
	Suppose that the GS algorithm is performed over payoffs $V : \cA \times \cA^+ \rightarrow \bbR$ and with $\cG$ as proposers. 
	Suppose further that $a' \neq a'' \implies V(a,a') \neq V(a,a'')$ for all $a \in \cA$. 
	Then, if the matching $\sM$ is returned by the GS algorithm, 
	$\sM \in S(V)$ is $\cG$-optimal and $(\cA \setminus \cG)$-pessimal. 
\end{lemmaApp}
\begin{proof}
	This result is well-known, cf. Theorem 3 given by \cite{roth2008deferred}.
\end{proof}

\section{PRELIMINARY RESULTS}\label{sec:app_prelim}

	Let $T(\cE) = \sum_{t=1}^T \mathbf{1}(\cE \text{ occurs at time step } t)$ denote the number of times event $\cE$ occurs over all $T$ time steps.

	\begin{lemmaApp}\label{lem:ucb_subgauss}
		Consider an agent faced with $K$ arms, as in the standard multi-armed bandit (MAB) problem.  
		Let $T \in \bbN_{>0}$ be the time horizon. 
		Suppose that, at each time step $t \in [T]$, 
		the agent is either: 
		(1) matched with arm $I_t$, in which case it receives reward $Y_{I_t,t}$; or 
		(2) unmatched, in which case it receives no reward. 
		Suppose $Y_{i,t}$ is drawn i.i.d. from a sub-Gaussian distribution with parameter $\sigma^2$ for all $i \in [K]$.
		Let $m(i) = \bbE[Y_{i,0}]$, $m^* = \max_{i \in [K]} m(i)$,
		and $T_{t}(i) = \sum_{\tau=1}^t \mathbf{1}(I_\tau = i)$. Let
		\begin{align*}
		v_t(i) = \frac{1}{T_{t-1}( i)} \sum_{t=1}^T Y_{i,t} \mathbf{1}(I_t = i) + \sqrt{\frac{2 \sigma^2 \alpha \log(t)}{T_{t-1}(i)}}  ,
		\end{align*}
		be the agent's upper confidence bound (UCB) for all arms $i \in [K]$, where $\alpha  > 2$. 
		Then,
		\begin{align*} 
			\bbE[T(I_t = i \cap v_t(i) \geq v_t(j) \cap m(i) < m(j) )] 
			&= \bbE \left[ \sum_{t = 1}^T \mathbf{1}( I_t = i \cap v_t(i) \geq v_t(j) \cap m(i) < m(j) ) \right]
			\\
			&\leq \frac{8 \sigma^2 \alpha \log(T)}{(m(j) - m(i) )^2} +  \frac{\alpha}{\alpha - 2} ,
		\end{align*}
		for all $i , j \in [K]$ and $i \neq j$, where the expectation is taken with respect to the randomness of rewards $Y_{I_t,t}$. 
	\end{lemmaApp}
	\begin{proof}
		This proof is adapted from the proof that upper bounds regret in the MAB problem under the UCB strategy, cf. Theorem 2.1 presented by \cite{bubeck2012regret}.

		Let  $\hat{m}_t(i) = \frac{1}{T_{t-1}( i)} \sum_{\tau=1}^t Y_{i,\tau} \mathbf{1}(I_\tau = i)$, 
		$\Delta(i,j) = | m(i) - m(j) |$, and 
		$\Psi^*(\epsilon) := \frac{\epsilon^2}{2 \sigma^2 }$. 
		If $I_t = i$, $v_t(i) \geq v_t(j)$, and $m(i) < m(j)$, then at least one of the following must be true: 
		\begin{align}
			&\hat{m}_t(j) + (\Psi^*)^{-1}\left(\frac{\alpha \log(t)}{T_{t-1}(j)} \right) \leq m(j) , \label{eq:UCB_1}
			\\
			&\hat{m}_t(i) > m(i) + (\Psi^*)^{-1}\left(\frac{\alpha \log(t)}{T_{t-1}(i)} \right) , \label{eq:UCB_2}
			\\
			&T_{t-1}(i) < \frac{\alpha \log(t)}{\Psi^*(\Delta(i,j) / 2)} \label{eq:UCB_3} .
		\end{align}
		To confirm that one of the prior statements must be true, suppose that all three statement are false. 
		Then, 
		\begin{align}
			v_t(j) &=
			\hat{m}_t(j) + (\Psi^*)^{-1}\left(\frac{\alpha \log(t)}{T_{t-1}(j)} \right) \label{eq:UCB_9}
			\\
			&> m(j) \label{eq:UCB_4}
			\\
			&= m(i) + \Delta(i,j)  \label{eq:UCB_5}
			\\
			&\geq m(i) + 2 (\Psi^*)^{-1}\left( \frac{\alpha \log(t)}{T_{t-1}(i)} \right) \label{eq:UCB_6}
			\\
			&\geq \hat{m}_t(i) + (\Psi^*)^{-1} \left( \frac{\alpha \log(t)}{T_{t-1}(i)} \right) \label{eq:UCB_7}
			\\
			&= v_t(i) , \label{eq:UCB_8}
		\end{align}
		where 
		\eqref{eq:UCB_9} and \eqref{eq:UCB_8} follow from the definition of $v_t$; 
		\eqref{eq:UCB_4} follows from the assumption that \eqref{eq:UCB_1} is false; 
		\eqref{eq:UCB_5} follows from the definition of $\Delta(i,j)$ and that $m(j) > m(i)$; 
		\eqref{eq:UCB_6} follows from the assumption that \eqref{eq:UCB_3} is false; 
		and 	\eqref{eq:UCB_7} follows from the assumption that \eqref{eq:UCB_2} is false. 
		However,  \eqref{eq:UCB_8} implies that $v_t(j) > v_t(i)$, which cannot be true by assumption. 
		
		Therefore, we have a contradiction, which implies that, if $I_t = i$, $v_t(i) \geq v_t(j)$, and $m(i) < m(j)$, then at least one of the statements \eqref{eq:UCB_1}-\eqref{eq:UCB_3} must be true. 
		Let $\bar{T}_{ij} = \lceil  \frac{\alpha \log(T)}{\Psi^*(\Delta(i,j) / 2)} \rceil$.
		Then,
		\begin{align}
			&\bbE[T(I_t = i \cap v_t(i) \geq v_t(j) \cap m(i) < m(j) )]  \nonumber
			\\
			&= \bbE \left[ \sum_{t = 1}^T \mathbf{1}( I_t = i \cap v_t(i) \geq v_t(j) \cap m(i) < m(j) ) \right] \nonumber
			\\
			&\leq \bbE \left[ \sum_{t = 1}^T \mathbf{1} \left( I_t = i \cap T_{t-1}(i) < \frac{\alpha \log(t)}{\Psi^*(\Delta(i,j) / 2)}  \right) \right] \nonumber
				\\
				& \qquad +  \bbE \Bigg[ \sum_{t=\bar{T}_{ij} +1}^T 
				\mathbf{1} \Bigg(
				\hat{m}_t(j) + (\Psi^*)^{-1}\left(\frac{\alpha \log(t)}{T_{t-1}(j)} \right)  \nonumber
				\\
				&\qquad \qquad 
					\leq m(j) 
					\cup 
					\hat{m}_t(i) > m(i) + (\Psi^*)^{-1}\left(\frac{\alpha \log(t)}{T_{t-1}(i)} \right) 
					\Bigg)
					\Bigg]
					\nonumber
			\\
			&\leq \bbE \left[ \sum_{t = 1}^T \mathbf{1} \left( I_t = i \cap T_{t-1}(i) < \frac{\alpha \log(T)}{\Psi^*(\Delta(i,j) / 2)}  \right) \right] \nonumber
			\\
			& \qquad +  \bbE \Bigg[ \sum_{t=\bar{T}_{ij} +1}^T 
			\mathbf{1} \Bigg(
			\hat{m}_t(j) + (\Psi^*)^{-1}\left(\frac{\alpha \log(t)}{T_{t-1}(j)} \right)  \leq m(j)  \nonumber
				\\
				& \qquad \qquad \qquad \qquad \qquad
				\cup 
				\hat{m}_t(i) > m(i) + (\Psi^*)^{-1}\left(\frac{\alpha \log(t)}{T_{t-1}(i)} \right) 
				\Bigg)
				\Bigg]
				\nonumber
			\\
				&\leq  \bar{T}_{ij}  + \bbE \Bigg[ \sum_{t=\bar{T}_{ij} +1}^T 
					\mathbf{1} \Bigg(
						\hat{m}_t(j) + (\Psi^*)^{-1}\left(\frac{\alpha \log(t)}{T_{t-1}(j)} \right) \leq m(j)  \nonumber
						\\
						&\qquad \qquad \qquad \qquad \qquad 
						\cup 
						\hat{m}_t(i) > m(i) + (\Psi^*)^{-1}\left(\frac{\alpha \log(t)}{T_{t-1}(i)} \right) 
					 \Bigg)
				  \Bigg] \nonumber
			\\
			&\leq  \bar{T}_{ij}  + \sum_{t=\bar{T}_{ij} +1}^T \Bigg[ 
				\bbP \Bigg(
				\hat{m}_t(j) + (\Psi^*)^{-1} \Bigg(\frac{\alpha \log(t)}{T_{t-1}(j)} \Bigg)  \leq m(j) 
				\Bigg)  \nonumber
				\\
					&\qquad \qquad \qquad \qquad \qquad
					+ 
					\bbP \Bigg( 
					\hat{m}_t(i) > m(i) + (\Psi^*)^{-1}\Bigg(\frac{\alpha \log(t)}{T_{t-1}(i)} \Bigg) 
					\Bigg)
				\Bigg] . \label{eq:UCB_expT}
		\end{align}
		It remains to bound the probabilities on the right-hand side in \eqref{eq:UCB_expT}. Beginning with the left term under the sum in \eqref{eq:UCB_expT}:
		\begin{align*}
			\bbP & \left(
			\hat{m}_t(j) + (\Psi^*)^{-1}\left(\frac{\alpha \log(t)}{T_{t-1}(j)} \right) \leq m(j) 
			\right) 
			\\
			&\leq 
				\bbP \left( \exists \tau \in \{1 , \hdots, t  \} :
				T_{t-1}(j) = \tau
				\cap 
				 \hat{m}_t(j) + (\Psi^*)^{-1} \left( \frac{\alpha \log(t) }{\tau} \right) \leq m(j)
				\right)
			\\
			&\leq \sum_{\tau=1}^t 
				\bbP \left(
				T_{t-1}(j) = \tau
				\cap 
				\hat{m}_t(j) + (\Psi^*)^{-1} \left( \frac{\alpha \log(t) }{\tau} \right) \leq m(j)
				\right)
			\\
			&\leq \sum_{\tau=1}^t \frac{1}{t^{\alpha}} %
			\\
			&= \frac{1}{t^{\alpha-1}} ,
		\end{align*}
		where the second-to-last line follows from the concentration properties of sub-Gaussian random variables \citep{vershynin2018high}.
		Incorporating the sum in \eqref{eq:UCB_expT},
		\begin{align}
			\sum_{t= \bar{T}_{ij} +1}^T \bbP  \left(
			\hat{m}_t(j) + (\Psi^*)^{-1}\left(\frac{\alpha \log(t)}{T_{t-1}(j)} \right) \leq m(j) 
			\right) 
			&\leq \sum_{t= \bar{T}_{ij} +1}^T \frac{1}{t^{\alpha-1}} \nonumber
			\\
				&\leq  \sum_{t= \bar{T}_{ij} +1}^\infty \frac{1}{t^{\alpha-1}} \nonumber
				\\
				&\leq  \sum_{t= 2}^\infty \frac{1}{t^{\alpha-1}} \nonumber
				\\
				&\leq  \int_{1}^\infty \frac{1}{t^{\alpha-1}} dt \nonumber
				\\
				&= \frac{1}{\alpha - 2} \label{eq:UCB_P1}.
		\end{align}
		Using a symmetric analysis, the right term under the sum in \eqref{eq:UCB_expT} can be upper bounded by the same quantity, such that
		\begin{align}
			\sum_{t=\bar{T}_{ij} +1}^T \bbP \left( 
			\hat{m}_t(i) > m(i) + (\Psi^*)^{-1}\left(\frac{\alpha \log(t)}{T_{t-1}(i)} \right) 
			\right)
			&\leq \frac{1}{\alpha - 2} . \label{eq:UCB_P2}
		\end{align}
		Combining with \eqref{eq:UCB_P1} and \eqref{eq:UCB_P2} with \eqref{eq:UCB_expT}: 
		\begin{align*}
			  \bbE[T(I_t = i \cap v_t(i) \geq v_t(j) \cap m(i) < m(j) )]  
			  &\leq 
			  \bar{T}_{ij} + \frac{2}{\alpha - 2}
			  \\
			  &\leq 
			  \frac{\alpha \log(T)}{\Psi^*(\Delta(i,j) / 2)} + 1 + \frac{2}{\alpha - 2}
			  \\
			  &\leq 
			  \frac{\alpha \log(T)}{\Psi^*(\Delta(i,j) / 2)} + \frac{\alpha }{\alpha - 2} , 
		\end{align*}
		which, after substituting in for the definition of $\Psi^*$, gives the result as stated in the lemma. 
		
		To connect this proof to Theorem 2.1 in \cite{bubeck2012regret}, note that $\Psi^*$ is the Legendre transform (convex-conjugate) of $Y - \bbE[Y]$, where $Y$ is a sub-Gaussian random variable with parameter $\sigma^2$ and that
		\begin{align*}
		(\Psi^*)^{-1} \left( \frac{\alpha \log(t)}{T_{t-1}(i)} \right)
		&= \sqrt{\frac{2 \sigma^2 \alpha \log(t)}{T_{t-1}(i)}} ,
		\end{align*}
		is the additive term in the UCB formula. 
	\end{proof}

	\begin{lemmaApp}\label{lem:cond_unstable}
		Consider the setup described in Section \ref{sec:setup}. 
		Let $\psi, \psi' : \cA \times \cA^+ \rightarrow \bbR$. 
		If $\ccM(\cdot \, ; \psi') \in S(V(\cdot, \cdot \, ; \psi'  ) )$ and $\ccM(\cdot \, ; \psi') \notin S(V(\cdot, \cdot \, ; \psi  ) )$, then there exists $a_1, a_2, a_3, a_4 \in \cA$ such that %
		$\ccM(a_1; \psi') = a_2$ and $\ccM(a_3; \psi') = a_4$, but
		$V(a_1,a_4 ; \psi) > V(a_1, a_2 ;\psi )$ and $V(a_4,a_1; \psi ) > V(a_4, a_3 ;\psi)$ and either $V(a_1,a_4 ; \psi') \leq V(a_1, a_2 ; \psi')$ or $V(a_4,a_1 ; \psi') \leq V(a_4, a_3 ; \psi')$.
	\end{lemmaApp}
	\begin{proof}
		If $\ccM(\cdot \, ; \psi') \notin S(V(\cdot, \cdot \, ; \psi ) )$, then there exists $a_1, a_2 \in \cA$ such that $\ccM(a_1; \psi') = a_2$ but $(a_1,a_2)$  is not stable under payoffs $V(\cdot, \cdot \, ; \psi  )$. 
		The latter implies that there exists $a_3, a_4 \in \cA$ such that $\ccM(a_3; \psi') = a_4$ and $V(a_1,a_4 ; \psi ) > V(a_1, a_2 ; \psi )$ and $V(a_4,a_1 ; \psi) > V(a_4, a_3 ; \psi)$. 
		However, because the matches $(a_1,a_2)$ and $(a_3,a_4)$ are stable under $V(\cdot, \cdot \, ; \psi' )$, either $V(a_1,a_4 ; \psi') \leq V(a_1, a_2 ; \psi')$ or $V(a_4,a_1 ; \psi') \leq V(a_4, a_3 ; \psi')$. 
	\end{proof}

\begin{lemmaApp}\label{lem:no_cost_transfer_T_bound}
	Consider the setup described in Section \ref{sec:setup}.
	Let $\Delta_{\min } = \min_{a_1 \in \cA,a_2 \in \cA^+,a_3 \in \cA^+ \setminus a_2} |\mu(a_1, a_2) - \mu(a_1, a_3)| $.
	Suppose there is no cost and no transfer such that $\cC = \cT = 0$. 
	Then, 
	\begin{align*}
	\bbE T(\ccM( \cdot \, ; \nu_t) \notin S(V(\cdot, \cdot, ; \mu) )  )   = \bbE  T(\ccM( \cdot \, ; \nu_t) \notin S(\mu)  )  &\leq 2 N^2 L    \left( \frac{8 \sigma^2 \alpha \log(T)}{ \Delta_{\min}^2} +  \frac{\alpha}{\alpha - 2} \right) .
	\end{align*} 
\end{lemmaApp}
\begin{proof}
	Under no cost and no transfer, $V(\cdot, \cdot \, ; \mu) = \mu(\cdot, \cdot)$ and $V(\cdot, \cdot \, ; \nu_t) = \nu_t(\cdot, \cdot)$. 
	Recall that, by construction, $\ccM(\cdot \, ; \nu_t) \in S(V(\cdot, \cdot \, ; \nu_t)) = S(\nu_t)$. 
	If, additionally, $\ccM(\cdot \, ; \nu_t) \notin S(V(\cdot, \cdot \, ; \mu)) = S(\mu)$, 
	then, by Lemma \ref{lem:cond_unstable}, there exists $a_1, a_2, a_3, a_4 \in \cA$ %
	such that $\ccM(a_1; \nu_t) = a_2$, $\ccM(a_3; \nu_t) = a_4$, $\mu(a_1,a_4 ) > \mu(a_1, a_2 )$, and $\mu(a_4,a_1 ) > \mu(a_4, a_3 )$ and either $\nu_t(a_1,a_4 ) \leq \nu_t(a_1, a_2)$ or $\nu_t(a_4,a_1 ) \leq \nu_t(a_4, a_3 )$.
	As such, 
	\begin{align*}
	\bbE  T & (\ccM( \cdot \, ; \nu_t)  \notin S(\mu)  ) 
	\\
	&\leq \bbE  T(\exists a_1, a_2, a_3, a_4 \in \cA : 
	\ccM(a_1; \nu_t) = a_2 \cap \ccM(a_3; \nu_t) = a_4
	\\
	& \qquad \cap \mu(a_1,a_4 ) > \mu(a_1, a_2 ) 
	\\
	& \qquad\cap \mu(a_4,a_1) > \mu(a_4, a_3 ) 
	\\
	& \qquad\cap (\nu_t(a_1,a_4 ) \leq \nu_t(a_1, a_2 )  \cup \nu_t(a_4,a_1 ) \leq \nu_t(a_4, a_3 ) ) 
	\\
	&\leq \bbE  T(\exists a_1, a_2, a_3, a_4 \in \cA : 	\ccM(a_1; \nu_t) = a_2 \cap \ccM(a_3; \nu_t) = a_4 
	\\
	& \qquad \qquad
	\cap \mu(a_1,a_4) > \mu(a_1, a_2)  
	\\
	& \qquad \qquad \cap \mu(a_4,a_1) > \mu(a_4, a_3 )  
	\\
	& \qquad \qquad \cap  \nu_t(a_1,a_4) \leq \nu_t(a_1, a_2 ) )
	\\
	& \qquad  + \bbE  T( \exists a_1, a_2, a_3, a_4 \in \cA :  \ccM(a_1; \nu_t) = a_2 \cap \ccM(a_3; \nu_t) = a_4 
	\\
	& \qquad \qquad \cap \mu(a_1,a_4) > \mu(a_1, a_2 )  
	\\
	& \qquad \qquad \cap \mu(a_4,a_1) > \mu(a_4, a_3 )  
	\\
	& \qquad \qquad \cap  \nu_t(a_4,a_1) \leq \nu_t(a_4, a_3 )  
	) 
	\\
	&\leq \bbE  T(\exists a_1, a_2, a_4 : \ccM(a_1; \nu_t) = a_2  \cap \mu(a_1,a_4 ) > \mu(a_1, a_2 ) 
	\cap \nu_t(a_1,a_4 ) \leq \nu_t(a_1, a_2 ) )
	\\ &\qquad 
	+  \bbE  T(\exists a_1, a_3, a_4 : 
	\ccM(a_3; \nu_t) = a_4  \cap \mu(a_4,a_1) > \mu(a_4, a_3 )  
	\cap  \nu_t(a_4,a_1 ) \leq \nu_t(a_4, a_3 )   )
	\\
	&= 2 \bbE  T(\exists a_1, a_2, a_3 :
	\ccM(a_1; \nu_t) = a_2 \cap
	\mu(a_1,a_3 ) > \mu(a_1, a_2 ) 
	\cap \nu_t(a_1,a_3 ) \leq \nu_t(a_1, a_2 ) ) . 
	\end{align*}
	Then, 
	\begin{align*}
	\bbE T & (\ccM( \cdot \, ; \nu_t)  \notin S(\mu)  )  
	\\
	&\leq 2 \bbE T(\exists a_1, a_2, a_3 :
	\ccM(a_1; \nu_t) = a_2 \cap
	\mu(a_1,a_3 ) > \mu(a_1, a_2 ) 
	\cap \nu_t(a_1,a_3 ) \leq \nu_t(a_1, a_2 ) ) 
	\\
	&\leq 2 \sum_{a_1, a_2, a_3 \in \cA } \bbE T( 	\ccM(a_1; \nu_t) = a_2 \cap  \mu(a_1,a_3 ) > \mu(a_1, a_2 ) 
	\cap \nu_t(a_1,a_3 ) \leq \nu_t(a_1, a_2 ) )
	\\
	&\leq 2 N^2 L    \left( \frac{8 \sigma^2 \alpha \log(T)}{ \Delta_{\min }^2} +  \frac{\alpha}{\alpha - 2} \right), 
	\end{align*} 
	where the last line follows from Lemma \ref{lem:ucb_subgauss} since, from the perspective of agent $a_1$, being matched with $a_2$ is identical to $a_1$ sampling arm $a_2$ in the MAB problem.
\end{proof}

	\section{PROOFS OF MAIN RESULTS (SECTION \ref{sec:results})}
\label{sec:app_no_CT}

	Recall that
	\begin{align*}
		\Delta_{\max} (a) &=   \max_{a_1,a_2 \in \cA^+} (\mu(a , a_1 ) ,
		- \mu(a , a_2 ) )
		\\
		\Delta_{\min} &= \min_{a_1 \in \cA, a_2 \in \cA^+, a_3 \in \cA^+ \setminus a_2} |\mu(a_1, a_2) - \mu(a_1, a_3)| .
	\end{align*}

\begin{proposition} 
	Suppose that there are no costs or transfers such that:
	$\cC(a_1,a_2 ; \psi) = 0$ and $\cT(a_1,a_2; \psi) = 0$ %
	for all $a_1 \in \cA$, $a_2 \in \cA^+$, and $\psi: \cA \times \cA^+ \rightarrow \bbR$.  
	If the GS algorithm is applied over $V(\cdot, \cdot \, ; \nu_t)$ at every $t \in [T]$, then the system is stable. 
	Moreover, 
	$
	\uR(a ; \ccM) \leq 2 N^2 L  \Delta_{\max} (a)   \left( \frac{8\sigma^2 \alpha \log(T)}{\Delta_{\min}^2} + \frac{\alpha}{\alpha - 2} \right)
	$
	for all $a \in \cA$. 
	However, under stability, it is not possible to guarantee fairness or high social welfare, and there exist settings $(\cA, \mu)$ for which $\bR(a ; \ccM) = \Omega(T)$ for at least one agent $a \in \cA$. 
\end{proposition}
	\begin{proof}
		By Lemma \ref{lem:GS_defection_proof}, the system is stable since the GS algorithm is applied at each time step. 
		Next, we perform the regret analysis. 
		
		\noindent \textbf{Logarithmic pessimal regret}. Recall that $\ucM_t^a \in  \arg \min_{\sM \in S(V(\cdot, \cdot \, ; \mu )) } \bbE \left[ U_t(a , \sM(a) ; \nu_t ) \right]$.
		Under no cost or transfer, $\bbE[ U_t(a, a' ; \nu_t)] = \mu(a,a')$ and $V(\cdot \, ; \cdot \, ; \mu) = \mu(\cdot, \cdot)$. 
		As such, 
		$\ucM_t^a \in \arg \min_{\sM \in S(\mu) } \mu(a, \sM(a) ) $, and the pessimal regret is 
		\begin{align*}
			\uR(a ; \ccM) &= \bbE \left[ \sum_{t=1}^T \left( 
			X_t(a , \ucM^a_t(a) ) - 	X_t(a  , \ccM(a ;  \nu_t ) )  
			\right) \right] 
			\\
			&= \sum_{t = 1}^T \mu(a , \ucM^a_t(a) ) 
				- \mu(a , \ccM(a ; \nu_t ) )
			\\
			&= \sum_{t = 1}^T \mathbf{1}(\ccM(\cdot \, ; \nu_t) \in S( \mu ) )\left( \mu(a , \ucM^a_t(a) ) 
			- \mu(a , \ccM(a ; \nu_t )) \right)
			\\
			& \qquad + \sum_{t = 1}^T \mathbf{1}(\ccM(\cdot \, ; \nu_t) \notin S(\mu  ) ) \left( \mu(a , \ucM^a_t(a) ) 
				- \mu(a , \ccM(a ; \nu_t ) ) \right)
			\\
			&\leq   \sum_{t = 1}^T \mathbf{1}(\ccM(\cdot \, ; \nu_t) \notin S(\mu  ) ) \left( \mu(a , \ucM^a_t(a) ) 
			- \mu(a , \ccM(a ; \nu_t ) ) \right), 
		\end{align*} 
		where the last line follows from the definition of $\ucM_t^a $. 
		Converting the sum over time steps to a sum over possible matches: 
		\begin{align*}
		\uR(a ; \ccM) &\leq   \sum_{t = 1}^T \mathbf{1}(\ccM(\cdot \, ; \nu_t) \notin S(\mu  ) ) \left( \mu(a , \ucM^a_t(a) ) 
		- \mu(a , \ccM(a ; \nu_t ) ) \right)
		\\
		&= \sum_{\sM' \notin S( \mu ) } T(\ccM(\cdot \, ; \nu_t) = \sM)
			\left( \mu(a , \ucM^a_t(a) ) 
			- \mu(a , \sM (a ; \nu_t) ) \right) .
		\end{align*} 
		Note that, in this setting, $\ucM^a_t(a)$ does not depend on $t$ because $\ucM_t^a \in \arg \min_{\sM \in S(\mu) } \mu(a, \sM(a) ) $. 
		Let $\ushort{\Delta}(a,a') =   \mu(a , \ucM^a_t(a) ) 
		- \mu(a , a' ) $. 
		Then,
		\begin{align*}
			\uR(a ; \ccM) &\leq  \sum_{\sM' \notin S( \mu ) } T(\ccM(\cdot \, ; \nu_t) = \sM)
		\ushort{\Delta}(a, \sM (a) ) 
		\\
			&\leq  \max_{a' \in \cA^+} \ushort{\Delta}(a,a') T(\ccM( \cdot \, ; \nu_t) \notin S(\mu) ) .
		\end{align*} 
		Then, applying Lemma \ref{lem:no_cost_transfer_T_bound}:
		\begin{align*}
		\uR(a ; \ccM) &\leq  \max_{a' \in \cA} \ushort{\Delta}(a,a') T(\ccM( \cdot \, ; \nu_t) \notin S(\mu) ) 
		\\
		&\leq 2 N^2 L  \Delta_{\max} (a)  \left( \frac{8 \sigma^2 \alpha \log(T)}{ \Delta_{\min }^2} +  \frac{\alpha}{\alpha - 2} \right) .
		\end{align*} 

		\noindent \textbf{Linear optimal regret}.
		Recall that 	$\bcM^a_t \in \arg \max_{\sM \in S(V(\cdot, \cdot \, ; \mu  )) } \bbE \left[ U_t(a , \sM(a) ; \nu_t ) \right]$.
		In the no-cost, no-transfer setting, $\bcM_t^a \in  \arg \max_{\sM \in S(\mu )} \mu(a, \sM(a) ))$ such that
		\begin{align}
			\bR(a ; \ccM) 
			&= \bbE \left[ \sum_{t=1}^T \left( 
			X_t(a , \bcM^a_t(a) ) - 	X_t(a  , \ccM(a ;  \nu_t ) )  
			\right) \right] 
			 \nonumber
			\\
			&= \sum_{t = 1}^T \mu(a , \bcM^a_t(a) ) 
			- \mu(a , \ccM(a ; \nu_t ) ) \nonumber
			\\
			&\leq T \Delta_{\max} (a) . \label{eq:optimal_reg_lin_T}
		\end{align}
		Since $a' \neq a'' \implies \mu(a,a') \neq \mu(a,a'')$ for all $a \in \cA$, $\Delta_{\max} (a) > 0$, which implies that $\bR(a ; \ccM)  = O(T)$ for all agents $a \in \cA$. 
		In other words, the optimal regret of any agent can be upper bounded by a function that grows linearly in $T$. 
		To show that this upper bound is tight, we now construct an example in which at least one agent $a \in \cA$ experiences regret that grows $\Omega(T)$, i.e., the optimal regret of at least one agent $a$ can be lower bounded by a function that grows linearly in $T$ such that $\bR(a ; \ccM) = \Omega(T)$.

		Suppose $\cU = \{u_1, u_2, u_3\}$, $\cP = \{p_1, p_2, p_3\}$, and
		\begin{align*}
			u_1 \succ_{\mu(p_1, \cdot)} u_2 \succ_{\mu(p_1, \cdot)}  u_3 ,& \qquad \qquad p_3 \succ_{\mu(u_1, \cdot)} p_2 \succ_{\mu(u_1, \cdot)}  p_1 ,
			\\
			u_1 \succ_{\mu(p_2, \cdot)} u_2 \succ_{\mu(p_2, \cdot)}  u_3 ,& \qquad \qquad p_3 \succ_{\mu(u_2, \cdot)} p_1 \succ_{\mu(u_2, \cdot)}  p_2 ,
			\\
			u_3 \succ_{\mu(p_3, \cdot)} u_2 \succ_{\mu(p_3, \cdot)}  u_1 ,& \qquad \qquad p_1 \succ_{\mu(u_3, \cdot)} p_2 \succ_{\mu(u_3, \cdot)}  p_3  .
		\end{align*}
		Then, the only two stable matchings under $V(\cdot, \cdot \, ; \mu)$ are
		\begin{align*}
			\sM_1(a) &= \begin{cases}
				p_2 ,& a = u_1 ,
				\\
				p_1 ,& a = u_2,
				\\
				p_3 ,& a = u_3,
				\\
				u_2 ,& a = p_1 ,
				\\
				u_1 ,& a = p_2,
				\\
				u_3 ,& a = p_3.
			\end{cases} 
			\qquad \qquad \qquad 
			\sM_2(a)~=~\begin{cases}
				p_2, & a = u_1 ,
				\\
				p_3 , & a = u_2,
				\\
				p_1 , & a = u_3,
				\\
				u_3, & a = p_1,
				\\
				u_1, & a = p_2,
				\\
				u_2, & a = p_3.
			\end{cases} 
		\end{align*}
		As such, $\bcM^{u_1}_t \in \{ \sM_1 , \sM_2 \}$, 
		$\bcM^{u_2}_t  = \sM_2$, 
		$\bcM^{u_3}_t  = \sM_2$, 
		$\bcM^{p_1}_t  = \sM_1$, 
		$\bcM^{p_2}_t  \in \{ \sM_1 , \sM_2 \}$, and 
		$\bcM^{p_3}_t  = \sM_1$. 
		The optimal regret of $p_1$ and $u_2$ can be lower bounded as follows. 
		\begin{align}
				\bR(a ; \ccM) 
				&= \bbE \left[ \sum_{t=1}^T \left( 
			X_t(a , \bcM^a_t(a) ) - 	X_t(a  , \ccM(a ;  \nu_t ) )  
			\right) \right] \nonumber
			\\
			&= \sum_{t = 1}^T \mu(a , \bcM^a_t(a) ) 
			- \mu(a , \ccM(a ; \nu_t ) ) \nonumber
				\\
			&\geq  T(\ccM(a ; \nu_t )  = \sM_1)
				 (\mu(a , \bcM^a_t(a) ) 
					- \mu(a , \sM_1(a))) \nonumber
						\\
						&\qquad + T(\ccM(a ; \nu_t )  = \sM_2)
						(\mu(a , \bcM^a_t(a) ) 
						- \mu(a , \sM_2(a))) , \label{eq:no_CT_eq_1}
		\end{align}
		which implies that
		\begin{align}
		\bR(p_1 ; \ccM) &\geq  T(\ccM( \cdot \, ; \nu_t)  = \sM_1)
		(\mu(p_1 ,\bcM^{p_1}_t (p_1) ) 
		- \mu(p_1 , \sM_1(p_1))) \nonumber
			\\
			&\qquad + T(\ccM( \cdot \, ; \nu_t)  = \sM_2)
			(\mu(p_1, \bcM^{p_1}_t (p_1) ) 
			- \mu(p_1 , \sM_2(p_1))) \nonumber
		\\
		&=  T(\ccM( \cdot \, ; \nu_t)  = \sM_1)
		(\mu(p_1 , \sM_1 (p_1) ) 
		- \mu(p_1 , \sM_1(p_1))) \nonumber
			\\
			&\qquad + T(\ccM( \cdot \, ; \nu_t)  = \sM_2)
			(\mu(p_1, \sM_1 (p_1) ) 
			- \mu(p_1 , \sM_2(p_1))) \nonumber
		\\
		&=  T(\ccM( \cdot \, ; \nu_t)  = \sM_2)
		(\mu(p_1, \sM_1(p_1) ) 
		- \mu(p_1 , \sM_2(p_1))) \nonumber
		\\
		&=  T(\ccM( \cdot \, ; \nu_t)  = \sM_2)
		(\mu(p_1, u_2 ) 
		- \mu(p_1 , u_3)) . \label{eq:max_reg_counterexample_1}
		\end{align}
		Recall that $\mu(p_1, u_2 ) 
		- \mu(p_1 , u_3) > 0$. 
		Therefore, the optimal regret for $p_1$ grows linearly in $T$ if $T(\ccM( \cdot \, ; \nu_t)  = \sM_2)$ grows linearly in $T$. 
		Similarly, applying \eqref{eq:no_CT_eq_1} to agent $u_2$ gives
		\begin{align}
			\bR(u_2 ; \ccM) &\geq  T(\ccM( \cdot \, ; \nu_t)  = \sM_1)
			(\mu(u_2 ,\bcM^{u_2}_t (u_2) ) 
			- \mu(u_2 , \sM_1(u_2))) \nonumber
			\\
				&\qquad + T(\ccM( \cdot \, ; \nu_t)  = \sM_2)
				(\mu(u_2, \bcM^{u_2}_t (u_2) ) 
				- \mu(u_2 , \sM_2(u_2))) \nonumber
			\\
			&=  T(\ccM( \cdot \, ; \nu_t)  = \sM_1)
			(\mu(u_2 , \sM_2 (u_2) ) 
			- \mu(u_2 , \sM_1(u_2))) \nonumber
				\\
				&\qquad + T(\ccM( \cdot \, ; \nu_t)  = \sM_2)
				(\mu(u_2, \sM_2 (u_2) ) 
				- \mu(u_2 , \sM_2(u_2))) \nonumber
			\\
			&=  T(\ccM( \cdot \, ; \nu_t)  = \sM_1)
			(\mu(u_2, \sM_2(u_2) ) 
			- \mu(u_2 , \sM_1(u_2))) \nonumber
			\\
			&=  T(\ccM( \cdot \, ; \nu_t)  = \sM_1)
			(\mu(u_2, p_3 ) 
			- \mu(u_2 , p_1 )) ) . \label{eq:max_reg_counterexample_2}
		\end{align}
		Recall that $\mu(u_2, p_3 ) 
		- \mu(u_2 , p_1 ) > 0$. Therefore, the optimal regret for $u_2$ grows linearly in $T$ if $T(\ccM( \cdot \, ; \nu_t)  = \sM_1)$ grows linearly in $T$. 
		
		We now show that either $T(\ccM( \cdot \, ; \nu_t)  = \sM_1)$ or $T(\ccM( \cdot \, ; \nu_t)  = \sM_2)$ grows linearly in $T$, or both do. 
		\begin{align*}
			T &= \sum_{\sM \in \cW} T(\ccM(\cdot \, ; \nu_t) = \sM)
			\\
			&= \sum_{\sM \in S(\mu)} T(\ccM(\cdot \, ; \nu_t) = \sM)
				+ \sum_{\sM \notin S(\mu)} T(\ccM(\cdot \, ; \nu_t) = \sM) , 
			\\
			\implies  \sum_{\sM \in S(\mu)} T(\ccM(\cdot \, ; \nu_t) = \sM) &= T -  \sum_{\sM \notin S(\mu)} T(\ccM(\cdot \, ; \nu_t) = \sM)
			\\
			&= T -  T(\ccM(\cdot \, ; \nu_t) \notin S(\mu))
			 \\
			 &\geq T - 2 N^2 L    \left( \frac{8 \sigma^2 \alpha \log(T)}{ \Delta_{\min}^2} +  \frac{\alpha}{\alpha - 2} \right) 
		\end{align*}
		which implies that $T(\ccM(\cdot \, ; \nu_t) = \sM_1) + T(\ccM( \cdot \, ; \nu_t) = \sM_2) = \Omega(T)$. 
		Since $T(\cE) \geq 0$ and $T > 0$, it is not possible that $T(\ccM( \cdot \, ; \nu_t) = \sM_1)$ and $T(\ccM(\cdot \, ; \nu_t)= \sM_2)$ both grow sub-linearly in $T$.
		Therefore, $T(\ccM( \cdot \, ; \nu_t) = \sM_1) = \Omega(T)$, $T(\ccM(\cdot \, ; \nu_t)= \sM_2) = \Omega(T)$, or both. 
		By \eqref{eq:max_reg_counterexample_1} and \eqref{eq:max_reg_counterexample_2}, the optimal regret of at least one agent grows $\Omega(T)$. 
		
		Intuitively, 
		since it is impossible for the matching at a given time step to simultaneously be both $\sM_1$ and $\sM_2$,
		at least one agent must incur a non-zero optimal regret for $\Omega(T)$ time steps. 

		\noindent \textbf{Fairness not guaranteed}. 
		To show that fairness is not guaranteed, it suffices to show the existence of an example in which the matchings are unfair. 
		That fairness is not guaranteed follows directly from the example under ``Linear optimal regret'' above. 
		In particular, we use the same example and show that it is possible for at least one agent's optimal regret to grow $O(\log(T))$ while another agent's regret grows $\Omega(T)$. 
		
		Suppose $T(\ccM( \cdot \, ; \nu_t) = \sM_2) = O(\log(T))$ such that $T(\ccM( \cdot \, ; \nu_t) = \sM_1) = \Omega(T)$.
		(This is possible under Algorithm \ref{alg:GS}, which returns matchings $\ccM( \cdot \, ; \nu_t)$ that are $\cP$-optimal at every time step by Lemma \ref{lem:GS_proposer_optimal}.)
		As such, for all $p \in \cP$,
		\begin{align*}
			\bR(p ; \ccM) &= \sum_{t=1}^T \mu(p , \bcM^p_t(p) ) 
			- \mu(p , \ccM(p ; \nu_t) )
			\\
			&= (\mu(p , \bcM^p_t(p) ) 
					- \mu(p , \sM_1(p ) )) T(\ccM( \cdot \, ; \nu_t) = \sM_1) 
					\\
				& \qquad \qquad 
				+ (\mu(p , \bcM^p_t(p) ) 
					- \mu(p , \sM_2(p ) )) T(\ccM( \cdot \, ; \nu_t) = \sM_2)
				\\
				&\qquad \qquad +\sum_{\sM \in \cW \setminus \{ \sM_1, \sM_2 \}} (\mu(p , \bcM^p_t(p) ) 
					- \mu(p , \sM(p ) )) T(\ccM( \cdot \, ; \nu_t) = \sM)
			\\
			&= (\mu(p , \bcM^p_t(p) ) 
				- \mu(p , \sM_2(p ) )) T(\ccM( \cdot \, ; \nu_t) = \sM_2)
				\\
				&\qquad \qquad +\sum_{\sM \in \cW \setminus \{ \sM_1, \sM_2 \}} (\mu(p , \bcM^p_t(p) ) 
				- \mu(p , \sM(p ) )) T(\ccM( \cdot \, ; \nu_t) = \sM)
			\\ &\leq \Delta_{\max}(a) \left[  T(\ccM( \cdot \, ; \nu_t) = \sM_2) + \sum_{\sM \in \cW \setminus \{ \sM_1, \sM_2 \}}  T(\ccM( \cdot \, ; \nu_t) = \sM) \right]
			\\
			&= O(\log(T)) ,
		\end{align*}
		where the second to last line follows from the fact that $\bcM^p_t = \sM_1$ for all $p \in \cP$, and the last line follows from Lemma \ref{lem:no_cost_transfer_T_bound} and that $\ccM$ can be chosen such that $T(\ccM( \cdot \, ; \nu_t) = \sM_2) = O(\log(T))$. 
		However, as shown in the example under ``Linear optimal regret'', when  $T(\ccM( \cdot \, ; \nu_t) = \sM_2) = O(\log(T))$, it must hold true that $T(\ccM( \cdot \, ; \nu_t) = \sM_1) = \Omega(T)$, which implies that the regret of a user must grow $\Omega(T)$. 
		By the definition of fairness (Definition \ref{def:fairness}), since there exists at least one agent (i.e., a provider) whose optimal regret grows $O(\log(T))$ and at least one agent (i.e., a user) whose optimal regret grows $\Omega(T)$, fairness is not guaranteed.

		\noindent \textbf{High social welfare not guaranteed}. 
		To show that high social welfare is not guaranteed, it suffices to show the existence of an example in which the social welfare is low. 
		Let us again use the example under ``Linear optimal regret''.
		Fix $\kappa > 1$. 
		Suppose $\ccM( \cdot \, ; \nu_t) = \sM_1$ and $\nu_t(u_2, p_3) > \frac{1}{\kappa} W_t(\ccM) =  \frac{1}{\kappa} \sum_{a \in \cA} \nu_t(a, \sM_1(a))$. 
		This outcome is possible because (1) $u_2$ and $p_3$ are not matched under $\sM_1$, which means that the values over which $W_t(\ccM)$ is summed does not include $\nu_t(u_2,p_3)$; 
		(2) the magnitude of $u_2$'s transient preferences does not affect the GS algorithm; and (3) it is possible for $\nu_t(u_2, p_3) > \nu_t(u_2,p_1)$ even when $\ccM(u_2 ; \nu_t) = p_1$ (simply apply the same logic as that given in the example under ``Linear optimal regret''). 
		As such, $\max_{\sM \in \cW} \sum_{a \in \cA} \nu_t(a, \sM(a)) \geq \nu_t(u_2, p_3) >  \frac{1}{\kappa} W_t(\ccM)$. 
		Since this reasoning holds true for any choice of $\kappa > 1$, by Definition \ref{def:SW}, the social welfare is low. 
		Therefore,  high social welfare is not guaranteed under no costs and no transfers.
	\end{proof}

\begin{theorem} 
	There exist cost and transfer rules $\cC(\cdot , \cdot \, ; \nu_t)$ and $\cT(\cdot , \cdot \, ; \nu_t)$ such that, 
	if the GS algorithm is applied over $V(\cdot, \cdot \, ; \nu_t)$ at every $t \in [T]$,
	then stability, low regret, 
	fairness, and $\frac{1}{2}$-high social welfare are guaranteed for all $(\cA, \mu)$.
\end{theorem}

\begin{proof}
	This is an existence statement. 
	We prove existence in the proof of Theorem \ref{thm:balanced_transfer}.
\end{proof}

	\section{PROOFS FOR PROPORTIONAL-COST, NO-TRANSFER SETTING (SECTION \ref{sec:prop_costs})}\label{sec:app_prop_cost}
	
	\begin{lemmaApp} \label{lem:prop_cost_all_stable}
		Suppose that $\cC$ and $\cT$ are set according to
		\eqref{eq:P_1}.
		If $\gamma  = 1$, then all matchings are stable, i.e., for any $\psi: \cA \times \cA^+ \rightarrow \bbR$ and $\sM \in \cW$, 
		$\sM \in S(V(\cdot, \cdot \, ; \psi))$.
	\end{lemmaApp}
	\begin{proof}
		When $\gamma = 1$, $V_t(\cdot, \cdot \, ; \psi)  = 0$ for all $\psi: \cA \times \cA^+ \rightarrow \bbR$. 
		Under any matching $\sM \in \cW$, there cannot exist another matching $\sM' \in \cW$ such that $V(a, \sM'(a) ; \psi) > V(a, \sM(a) ; \psi)$ for some $a$ since $0 > 0$ is always false, which implies that $\sM$ is stable. This result holds for any $\sM \in \cW$.
	\end{proof}

	\begin{proposition} 
		Suppose that $\cC$ and $\cT$ are set according to
		\eqref{eq:P_1}.
		If the GS algorithm is applied over $V(\cdot, \cdot \, ; \nu_t)$ at every $t \in [T]$, then the system is stable. 
		If $\gamma \in [0,1)$, 
		$\uR (a ; \ccM) \leq 2 N^2 L (1 - \gamma)  \Delta_{\max} (a)   \left( \frac{8\sigma^2 \alpha \log(T)}{(1 - \gamma)^2 \Delta_{\min}^2} + \frac{\alpha}{\alpha - 2} \right)$
		and, if $\gamma = 1$, 
		$\uR(a ; \ccM) \leq 0$.
		However, under stability, it is not possible to guarantee fairness or high social welfare, and there exist settings $(\cA, \mu)$ for which $\bR(a ; \ccM) = \Omega(T)$ for at least one agent $a \in \cA$. 
	\end{proposition}

	\begin{proof}
		For $\gamma \in [0,1)$, the setting described in this claim is equivalent to that in Proposition \ref{prop:no_P_no_T}, except that the payoffs are scaled by $(1 - \gamma)$.
		To see this, observe that $V(a, a'; \psi) = \psi(a,a') - \gamma \psi(a,a') = (1 - \gamma) \psi(a,a')$. 
		Therefore, the same results as in Proposition \ref{prop:no_P_no_T} can be applied here with the appropriate scaling. 
		
		We consider $\gamma = 1$ next. 
		By Lemma \ref{lem:GS_defection_proof}, the system is stable since the GS algorithm is applied at each time step. 
		Next, we perform the regret analysis. 
		
		\noindent \textbf{Non-positive pessimal regret}.
		By \eqref{eq:P_1},  $\cT = 0$ and $\cC(a, a' ; \nu_t)  = \nu_t(a, a')$ for all $a \in \cA$, $a' \in \cA^+$, and $t \in [T]$ when $\gamma = 1$. 
		As such, the observed payoff at each time step $t$ is $U_t(a, a' ; \nu_t) = X_t(a, a') - \nu_t(a, a')$, and $\bbE[U_t(a,a' ; \nu_t)] = \mu(a,a') - \bbE[\nu_t(a,a')]$. 
		Recall that
		\begin{align*}
			\ucM_t^a &\in \argmin_{\sM \in S(V(\cdot, \cdot \, ; \mu ) ) } \bbE[U_t(a, \sM(a) ; \nu_t)]
			\\
			&= \argmin_{\sM \in S(V(\cdot, \cdot \, ; \mu) ) } \bbE[X_t(a, \sM(a)) - \nu_t(a, \sM(a)) ] .
		\end{align*}
		Under the cost rule, $V(\cdot, \cdot \, ; \psi) = 0$ for all choices of $\psi$. By Lemma \ref{lem:prop_cost_all_stable},
		\begin{align*}
		\ucM_t^a &\in \argmin_{\sM \in \cW } \bbE[X_t(a, \sM(a)) - \nu_t(a, \sM(a)) ] 
		\\
		&= \argmin_{\sM \in \cW } \left(\mu(a, \sM(a) ) - \bbE [ \nu_t(a, \sM(a))] \right) .
		\end{align*}
		The pessimal regret for $a$ is
		\begin{align}
		\uR(a ; \ccM) &= \sum_{t=1}^T \bbE\left[ U_t(a, \ucM(a)_t^a ; \nu_t) - U_t(a , \ccM(a ; \nu_t); \nu_t) \right] \nonumber
		\\
		&= \sum_{t=1}^T \bbE\left[ X_t(a,\ucM_t^a (a) ) - \nu_t(a, \ucM_t^a (a) ; \nu_t) - X_t(a , \ccM(a ; \nu_t ) ) + \nu_t(a ; \ccM(a ; \nu_t ) ; \nu_t) \right] \nonumber
		\\
		&= \sum_{t=1}^T \bbE\left[ X_t(a,\ucM_t^a (a) ) - \nu_t(a, \ucM_t^a (a) ; \nu_t) - X_t(a , \ccM(a ; \nu_t ) ) + \nu_t(a ; \ccM(a ; \nu_t ) ; \nu_t) \right] . \nonumber
		\end{align}
		Recall that both domains $S(V(\cdot, \cdot \, ; \mu))$ and $S(V(\cdot, \cdot \, ; \nu_t))$ over which $\ucM_t^a$ and $\ccM( \cdot \, ; \nu_t)$ are chosen are unrestricted (i.e., $S(V(\cdot, \cdot \, ; \mu)) = S(V(\cdot, \cdot \, ; \nu_t)) = \cW$). 
		Then, by definition of $\ucM_t^a$,
		\begin{align*}
		\bbE&\left[ X_t(a,\ucM_t^a (a) ) - \nu_t(a, \ucM_t^a (a) ) - X_t(a , \ccM(a ; \nu_t ) ) + \nu_t(a ; \ccM(a ; \nu_t ) ) \right] 
		\leq 0 ,
		\end{align*}
		which implies $\uR(a ; \ccM) \leq 0$, as stated in the result.

		\noindent \textbf{Linear optimal regret}. 
		By definition, the optimal regret of any agent grows $O(T)$ (see \eqref{eq:optimal_reg_lin_T} for reference), i.e., the optimal regret of any agent can be upper bounded by a function that grows linearly in $T$ such that $\bR(a ; \ccM) = O(T)$ for all $a \in \cA$. 
		To show that this upper bound is tight, we now construct an example in which at least one agent $a \in \cA$ experiences regret that grows $\Omega(T)$, i.e., the optimal regret of at least one agent $a$ can be lower bounded by a function that grows linearly in $T$ such that $\bR(a ; \ccM) = \Omega(T)$. 
		\begin{align}
			\bR(a ; \ccM) &= \sum_{t=1}^T \bbE\left[ U_t(a, \bcM_t^a(a) ; \nu_t) - U_t(a , \ccM(a ; \nu_t); \nu_t) \right] \nonumber
			\\
			&= \sum_{t=1}^T \bbE\left[ X_t(a,\bcM_t^a (a) ) - \nu_t(a, \bcM_t^a (a) ; \nu_t) - X_t(a , \ccM(a ; \nu_t ) ) + \nu_t(a ; \ccM(a ; \nu_t ) ; \nu_t) \right] \nonumber
			\\
			&= \sum_{t=1}^T - \bbE\left[ \sqrt{\frac{2 \sigma^2 \alpha \log(t)}{T_{t-1} (a, \bcM_t^a (a))}} \right] + \bbE\left[ \sqrt{\frac{2 \sigma^2 \alpha \log(t)}{T_{t-1} (a,  \ccM(a ; \nu_t ) )}} \right]  \nonumber
			\\
			&= \sum_{t=1}^T \sqrt{2 \sigma \alpha \log(t)}  \bbE\left[  \sqrt{\frac{1}{T_{t-1} (a,  \ccM(a ; \nu_t ) )}} -  \sqrt{\frac{1}{T_{t-1} (a, \bcM_t^a (a))}} \right] 
			 \label{eq:cost_no_transfer_eq1} .
		\end{align}
		Recall that
		\begin{align*}
			\bcM_t^a &\in \argmax_{\sM \in S(V(\cdot, \cdot \, ; \mu ) ) } \bbE[U_t(a, \sM(a) ; \nu_t)]
			\\
			&= \argmax_{\sM \in S(V(\cdot, \cdot \, ; \mu ) ) } \bbE[X_t(a, \sM(a)) - \nu_t(a, \sM(a)) ] 
			\\
			&= \argmax_{\sM \in S(V(\cdot, \cdot \, ; \mu ) ) } - \bbE \left[ \sqrt{\frac{1}{T_{t-1} (a, \sM (a))}}  \right]
			\\
			&= \argmax_{\sM \in S(V(\cdot, \cdot \, ; \mu ) ) } \bbE \left[ T_{t-1} (a, \sM (a))  \right]
			.
		\end{align*}
		Under the cost rule, $V(\cdot, \cdot \, ; \psi) = 0$ for all choices of $\psi$, which implies that
		\begin{align}
		\bcM_t^a &\in \argmax_{\sM \in \cW } \bbE \left[ T_{t-1} (a, \sM (a))  \right] . \label{eq:ex_2_opt_match}
		\end{align}
		Using this observation, we now construct an example to show that the optimal regret of an agent can grow linearly. 
		To do so, we make repeated use of Lemma \ref{lem:prop_cost_all_stable}, which implies that both of the domains $S(V(\cdot, \cdot \, ; \mu))$ and $S(V(\cdot, \cdot \, ; \nu_t))$ over which $\bcM_t^a$ and $\ccM( \cdot \, ; \nu_t)$ are chosen are unrestricted (i.e., $S(V(\cdot, \cdot \, ; \mu)) = S(V(\cdot, \cdot \, ; \nu_t)) = \cW$). 
		
		Suppose $p_1 , p_2 \in \cP$, where $N \geq L > 2$. 
		Suppose that, at every time step, $\ccM$ matches $p_1$ to $u_1$, $p_2$ uniformly at random to a user $u \in \cU \setminus u_1$,
		and all remaining providers arbitrarily to complete the matching. 
		Let $\eta^*_t(a) = \argmax_{a' \in \cA} \bbE \left[ T_{t-1}(a, a') )\right]$.
		Then,  by Lemma \ref{lem:prop_cost_all_stable}, such matchings are stable and can therefore be the matching chosen by the platform at any time step $t$ to satisfy stability. 
		Under this matching strategy, $p_1$ is guaranteed to be matched to its optimal match $\bcM_t^{p_1}(p_1)$ at any time step $t > \max_{u \in \cU} T_0(p_1,u)$, which implies  $\bR(p_1 ; \ccM) = O(1)$. 
		To see this, observe that, after matching $p_1$ with $u_1$ for $\max_{u \in \cU} T_0(p_1,u)$ times, $\eta^*_t(p_1) = u_1$. In other words, the agent with which $p_1$ has been most matched is $u_1$, which means that it is $u_1$ is $p_1$'s optimal match by \eqref{eq:ex_2_opt_match}. 
		
		On the other hand, whenever $|\eta_t^*(p_2)| \neq N$,  $p_2$ is not matched to a user $u \in \eta_t^*(p_2)$ and incurs non-zero regret with probability at least $1/N$.
		The event $|\eta_t^*(p_2)| = N$ can occur at most $\lceil T/N \rceil$ times because $|\eta_t^*(p_2)| = N$ implies that $p_2$ has sampled all users the same number of times at time step $t$. 
		As a result, $\bR(p_2 ; \ccM) = \Omega(T)$. 
		We have therefore shown that it  is not possible to guarantee low optimal regret (i.e., not possible to ensure that $\bR(a ; \ccM) = O(\log(T))$ for all $a \in \cA$) under $\gamma = 1$. 
		
		Intuitively, because all matchings are stable by Lemma \ref{lem:prop_cost_all_stable}, the platform is afforded so much flexibility that it can choose matchings with detrimental regret outcomes. 
		We now show that this flexibility also allows the platform to choose matchings that are unfair.
		
		\noindent \textbf{Fairness not guaranteed}. 
		That matchings are not guaranteed to be fair under proportional costs and $\gamma = 1$ follows directly from the example provided under ``Linear optimal regret'' above. 
		We showed that, in this example, there exists at least one agent $a$ for which $\bR(a ; \ccM) = O(1) = O(\log(T))$ while there exists at least one other agent $a'$  for which $\bR(a' ; \ccM) = \omega(\log(T)) = \Omega(T)$. 
		By Definition \ref{def:fairness}, this outcome is unfair, showing that fairness is not guaranteed under $\gamma = 1$. 
		
		\noindent \textbf{High social welfare not guaranteed}.
		Under the cost rule, $V(\cdot, \cdot \, ; \nu_t) = 0$ at all time steps $t \in [T]$. 
		Therefore, by Definition \ref{def:SW}, the social welfare at every time step is low. 
	\end{proof}

\begin{corollary} 
	Suppose that $\cC$ and $\cT$ are set according to \eqref{eq:P_1}
	and that the platform wishes to maximize  $F_t: \cW \rightarrow \bbR$ at every  $t \in [T]$. 
	Then, 
	when $\gamma = 1$, 
	there is no performance cost to stability: i.e., $\arg \max_{\sM \in \cW} F_t(\sM) \in S(V(\cdot, \cdot \, ; \nu_t))$ for all $t \in [T]$.
\end{corollary}
\begin{proof}
	This result follows from Lemma \ref{lem:prop_cost_all_stable}. Since all matchings are stable at every time step, all matchings are stable, including $\arg \max_{\sM \in \cW} F_t(\sM)$.
\end{proof}

	\section{PROOFS FOR BALANCED TRANSFER (SECTION \ref{sec:balanced_transfer})} \label{sec:app_balanced_transfer}

		Recall that the GS algorithm in Algorithm \ref{alg:GS} is performed with providers as the proposers (i.e., in the notation of Lemma \ref{lem:GS_proposer_optimal}, $\cG = \cP$). 
		A second version of the GS algorithm in which users are the proposers (i.e., $\cG = \cU)$ is given Algorithm \ref{alg:GS_user}.

		In our analysis, we use the following well-known result. 
		
		\begin{lemmaApp}\label{lem:unique_matching_proposer_opt}
			Suppose $V: \cA \times \cA^+ \rightarrow \bbR$ denotes a payoff function and $a' \neq a'' \implies V(a,a') \neq V(a,a'')$ for all $a \in \cA$. 
			Then, a matching $\sM \in \cW$ is the unique stable matching under payoffs $V$ (i.e., $\sM = S(V)$) if and only if the GS algorithm performed with providers as proposers and the GS algorithm performed with users as proposers return the same matching $\sM$.
		\end{lemmaApp}
		\begin{proof}
			Denote the matching returned when the GS algorithm is performed with providers as proposers by $\sM_1$ and the matching returned when the GS algorithm is performed with users as proposers by $\sM_2$. 
		
		To prove the if-and-only-if statement, let's begin with the forward direction: that $\sM$ being the unique stable matching implies $\sM_1 = \sM_2$. This result is trivial because $\sM_1$ and $\sM_2$ must be stable by Lemma \ref{lem:GS}. Therefore, since there is only one stable matching under $V$, $\sM_1 = \sM_2 = \sM$.

		{\centering
			\begin{minipage}[t]{.95\linewidth}
				\begin{algorithm}[H]
					\SetAlgoLined
					\KwIn{Set of agents $\cA = \cU \cup \cP$, where the set of users is  $\cU = \{u_1, u_2, \hdots, u_N \}$, the set of providers is $\cP  = \{ p_1, p_2, \hdots, p_L \}$, $\cU \cap \cP = \emptyset$, and $N \geq L$. A payoff function $V: \cA \times \cA^+ \rightarrow \bbR$.
					}
					\KwOut{Matching $\sM \in S(V) \subset \cW$. }
					\BlankLine
					Initialize matching $\sM: \cA \rightarrow \cA^+$ such that $\sM(a) = \emptyset$ for all $a \in \cA$\;
					Initialize empty (FIFO) queues $Q(u) = [ \hspace{1pt} ]$ for all $u \in \cU$\;
					\BlankLine
					\tcp{Fill each user's queue with providers in order of decreasing preference.}
					\For{$u \in \cU$}{
						\For{$i = 1, 2, \hdots, L$}{
							Append $r^{-1}(i ; V(u, \cdot))$ to $Q(u)$\tcp*[l]{Add $u$'s $i$-th ranked user.}
						}
					}
					
					\BlankLine
					\tcp{As long as there exists a user who is unmatched and has a non-empty queue...}
					\While{$\exists u \in \cU : \sM(u) = \emptyset \cap |Q(u)| > 0$}{\label{line:GS_pick_user}
						$p \leftarrow \text{pop}(Q(u))$\tcp*[l]{User $u$'s favorite provider of those remaining in $u$'s queue.}
						\BlankLine
						\tcp{If provider $p$ is unmatched, match $u$ and $p'$.}
						\uIf{$\sM(p) = \emptyset$}{
							$\sM(p) \leftarrow u$\;
							$\sM(u) \leftarrow p$\;
						}
						\BlankLine
						\tcp{If provider $p$ prefers $u$ to its current match $\sM(p)$, match $u$ and $p$.}
						\uElseIf{$V(p, u ) > V(p , \sM(p) )$}{
							$\sM'( \sM(p) ) = \emptyset$\;
							$\sM(p) \leftarrow u$\;
							$\sM(u) \leftarrow p$\;
						}
					}
					Return $\sM$\;
					\caption{Gale-Shapley algorithm (with users as proposers)} \label{alg:GS_user}
					\algorithmfootnote{In the GS algorithm presented here, the users are the proposers. The version of the algorithm in which providers are the proposers is given in Section \ref{sec:results}.}
				\end{algorithm}
			\end{minipage}
			\par
		}

		To show the backward direction, we must show that $\sM_1 = \sM_2 = \sM$ implies that $\sM$ is the unique stable matching under $V$.
		By Lemma \ref{lem:GS_proposer_optimal}, $\sM_1$ is $\cP$-optimal and $\cU$-pessimal, while $\sM_2$ is $\cU$-optimal and $\cP$-pessimal. 
		Since $\sM_1 = \sM_2 = \sM$,
		applying Definition \ref{def:pessimal_optimal} gives that:
		\begin{align*}
		V(p,\sM(p)) &\geq V(p,\sM'(p)) ,
		\\
		V(p,\sM(p)) &\leq V(p,\sM'(p)) ,
		\\
		V(u,\sM(u)) &\leq V(u,\sM'(u)) ,
		\\
		V(u,\sM(u)) &\geq V(u,\sM'(u)) ,
		\end{align*}
		for all $p \in \cP$, $u \in \cU$, and $\sM' \in S(V)$. 
		However, the first two statements imply that $V(p,\sM(p)) \geq V(p,\sM'(p)) \geq V(p,\sM(p)$ for all $p \in \cP$ and all stable matchings $\sM' \in S(V)$. 
		Similarly, the latter two statements imply that $V(u,\sM(u)) \leq V(u,\sM'(u)) \leq V(u,\sM(u))$  for all $u \in \cU$ and all stable matchings $\sM' \in S(V)$. 
		These outcomes are only possible if $\sM' = \sM$ for all $\sM' \in S(V)$, which implies that there is a unique stable matching $\sM = \sM_1 = \sM_2$. 
	\end{proof}

		\begin{definition}
			A function $h: \cA \times \cA^+ \rightarrow \bbR$ is \emph{pairwise-unique} if $h(a_1,a_2) \neq h(a_3, a_4)$ unless $(a_1,a_2) = (a_3, a_4)$ or $(a_2,a_1) = (a_3,a_4)$. 
		\end{definition}
		Recall that we assume pairwise-uniqueness in Section \ref{sec:setup}.
		In the Appendix, in the remainder of this section, 
		if pairwise-uniqueness is needed, it is mentioned explicitly. 
		Outside of this section, it is not needed, except for Theorem \ref{thm:possibility_result}.
		However, as noted in the main text, pairwise-uniqueness
		is not strictly necessary for any of the results. 
		Instead, the platform could have a consistent tie-breaking rule. 
		Specifically, the platform would first place matches (i.e., user-provider pairs) in some order. This order could be principled or arbitrary. 
		Then, whenever there is a tie, the algorithm breaks ties based on this order. 
		The same stability, regret, fairness, and social welfare results would hold with small modifications. 
		In particular, the set of stable matchings in this scenario should be restricted to stable matchings that can be obtained using this tie-breaking rule.

	{\centering
		\begin{minipage}{.95\linewidth}
			\begin{algorithm}[H]
				\SetAlgoLined
				\KwIn{Set of agents $\cA = \cU \cup \cP$, where the set of users is  $\cU = \{u_1, u_2, \hdots, u_N \}$, the set of providers is $\cP  = \{ p_1, p_2, \hdots, p_L \}$, $\cU \cap \cP = \emptyset$, and $N \geq L$. Preferences $\psi :  \cA \times \cA^+ \rightarrow \bbR$.
				}
				\KwOut{Matching $\sM \in S(V) \subset \cW$, where $V(a,a' ) = \frac{1}{2} (\psi(a,a') + \psi(a',a))$. }
				\BlankLine
				Initialize matching $\sM: \cA \rightarrow \cA^+$ such that $\sM(a) = \emptyset$ for all $a \in \cA$\;
				$w(e) \leftarrow \psi(u,p) + \psi(p,u)$ for all $e = (u,p)$, where $u \in \cU$ and $p \in \cP$\;
				Fill a (FIFO) queue $\bar{Q}$ with edges $e$ in decreasing order of weight $w(e)$\;
				
				\BlankLine
				\tcp{As long as there exists a provider who is unmatched...}
				\While{$\exists p' \in \cP : \sM(p') = \emptyset$}{ \label{line:greedy_pick_proposer}
					$e = (u,p) \leftarrow \text{pop}(\bar{Q})$\tcp*[l]{Edge with highest weight that remains in the queue.}
					\BlankLine
					\tcp{If edge $e$ does not conflict with an existing match, then add it to $\sM$}
					\If{$\sM(u) = \emptyset$ and $\sM(p) = \emptyset$}{
						$\sM(u) \leftarrow p$\;
						$\sM(p) \leftarrow u$\;
					}
				}
				Return $\sM$\;
				\caption{Greedy algorithm to determine a stable matching under the balanced transfer rule.} \label{alg:greedy_balanced}
			\end{algorithm}
		\end{minipage}
		\par
	}

	\begin{lemmaApp}\label{lem:pairwise_unique_matching}
		Suppose that $\cC$ and $\cT$ are set according to \eqref{eq:T_1}-\eqref{eq:T_2}.
		If payoffs $V(\cdot , \cdot \, ; \psi)$ are pairwise-unique, then there is a unique stable matching under these payoffs: i.e., $|S(V(\cdot, \cdot \, ; \psi))| =  1$. 
	\end{lemmaApp}
		\begin{proof}
			Under \eqref{eq:T_1}, $V(a,a' ; \psi) = \frac{1}{2} (\psi(a,a') + \psi(a',a)) = V(a', a ; \psi)$. 
			In other words, $V(\cdot, \cdot \, ; \psi)$ is symmetric. 
			Graphically, this means that the edges between users and providers are undirected. 
			If $V(\cdot, \cdot \, ; \psi)$ is pairwise-unique, then all edge weights in this undirected graph are unique. 
			The unique stable matching can be found using a greedy algorithm, as given in Algorithm \ref{alg:greedy_balanced}. 
			In this algorithm, all edges are first sorted from highest to lowest weight. 
			Then, beginning with the edges with the highest weights, the match $(u,p)$ associated with the given edge is added to the matching $\sM$ unless it conflicts with an existing match in $\sM$. 
			Repeat this process until all providers are matched such that $\sM(p) \neq \emptyset$ for all $p \in \cP$.

			To see why the resulting matching $\sM$ is stable, observe that it is identical to the GS algorithm in Algorithm \ref{alg:GS} as long as the unmatched provider $p$ who proposes in Line  \ref{line:pick_proposer} of Algorithm \ref{alg:GS} is always the unmatched provider with the greatest unproposed edge weight such that $\max_{u \in Q(p)} V(p, u ; \psi) \geq \max_{u \in Q(p')} V(p', u ; \psi)$ for all $p' \in \cP$ for which $\sM(p') = \emptyset$.
			This is equivalent to the greedy algorithm described above because it is not possible for any proposed match to override a previous one. As such, all proposals that conflict with an existing match are discarded. 
			As the GS algorithm is agnostic to the order of proposers, such an ordering is allowed, which implies that $\sM$ is stable by Lemma \ref{lem:GS}.
			
			To see why the resulting stable matching is unique, we make use of Lemma \ref{lem:unique_matching_proposer_opt}. 
			In particular, we have already shown that $\sM$ is the result of running the GS algorithm with providers as proposers. If we additionally show that $\sM$ can be obtained by running the GS algorithm with users as proposers, then $\sM$ is the unique stable matching under $V(\cdot, \cdot \, ; \psi)$. 
			We use the same observation as above. 
			$\sM$ can be obtained by running Algorithm \ref{alg:GS_user}, in which the user $u$ chosen to  propose in Line \ref{line:GS_pick_user} is the user with the greatest unproposed edge weight such that $\max_{p \in Q(u)} V(u, p ; \psi) \geq \max_{p \in Q(u')} V(u', p ; \psi)$ for all $u' \in \cU$ for which $\sM(u') = \emptyset$ and $|Q(u') | > 0$.
			However, this is precisely the same algorithm as that described in the paragraph directly above because $V(a, a' ; \psi) = V(a', a ; \psi)$ is symmetric. 
			Therefore, Algorithms \ref{alg:GS} and \ref{alg:GS_user} both return $\sM$, which implies that $\sM$ is the unique stable matching by Lemma \ref{lem:unique_matching_proposer_opt}. 
		\end{proof}
	
	Let $\rho(a,a') = \frac{1}{2}( \mu(a,a') + \mu(a',a))$ and
	$\Delta^{\rho}_{\min} = \min_{ a_1 \in \cA,a_2 \in \cA^+, a_3 \in \cA^+ \setminus a_2} | \rho(a_1,a_2) -  \rho(a_1,a_3) |$. 
	Note that, if $\rho$ is pairwise-unique, $S(\rho)$ contains only one element,
	which we denote by $\sM^{*,\rho} = S(\rho)$.
	Accordingly, let
	$\Delta_{\max}^{*,\rho}(a) = \max_{a' \in \cA^+} (\rho(a,\sM^{*,\rho}(a)) - \rho(a,a') )$.

\begin{theorem}
	Suppose that $\cC$ and $\cT$ are set according to \eqref{eq:T_1}-\eqref{eq:T_2}.
	Then, 
	$| S(\rho) | = 1$.
	Let $\sM^{*,\rho}$ be the only element in $S(\rho)$ 
	and
	$\Delta_{\max}^{*,\rho}(a) = \max_{a' \in \cA^+} (\rho(a,\sM^{*,\rho}(a)) - \rho(a,a') )$. 
	If the GS algorithm is applied over $V(\cdot, \cdot \, ; \nu_t)$ at every $t \in [T]$, then the system is stable, the social welfare is $\frac{1}{2}$-high at all $t \in [T]$,
	$\ccM$ is fair, and
	$\uR(a; \ccM) = \bR(a ; \ccM) \leq 
	\Delta_{\max}^{*,\rho}(a) N^2 L    \left( \frac{8 \sigma^2 \alpha \log(T)}{ (\Delta^{\rho}_{\min})^2} +  \frac{\alpha}{\alpha - 2} \right)$
	for all $a \in \cA$. 
\end{theorem}

	\begin{proof}
		By Lemma \ref{lem:GS_defection_proof}, the system is stable since the GS algorithm is applied at each time step. 
		Next, we perform the regret analysis. 
		
		\noindent \textbf{Logarithmic optimal regret}.
		By \eqref{eq:T_1}, $\cC = 0$ and $\cT(a, a' ; \nu_t)  = \frac{1}{2} (\nu_t(a', a) - \nu_t(a, a'))$ for all $a, a' \in \cA$ and $t \in [T]$.
		As such, the observed payoff at each time step $t$ is $U_t(a, a' ; \nu_t) = X_t(a, a') +  \frac{1}{2} (\nu_t(a', a) - \nu_t(a, a'))$, and
		\begin{align}
			\bbE[U_t(a,a' ; \nu_t)] 
				&= \mu(a,a')  + \frac{1}{2} \bbE[\nu_t(a', a) - \nu_t(a, a')]  \nonumber
				\\
				&= \frac{1}{2} ( \mu(a,a') +  \mu(a',a) ) 
				+ \bbE\left[ 
				\sqrt{\frac{2 \sigma^2 \alpha \log(T)}{T_{t-1} (a',a)}} 
				- 	\sqrt{\frac{2 \sigma^2 \alpha \log(T)}{T_{t-1} (a,a')} }
					 \right] \nonumber
				\\
				&= \frac{1}{2} ( \mu(a,a') +  \mu(a',a) ) ,\label{eq:transfer_no_cost_eq1}
		\end{align}
		where the last line follows from the fact that the number of times agent $a$ samples $a'$ is the same as the number of times agent $a'$ samples $a$ since samples are obtained from being matched, which means that $T_{t-1}(\cdot, \cdot)$ is therefore symmetric. 
		
		Recall that $\uR(a ; \ccM) \leq \bR(a ; \ccM)$. 
		We therefore focus on upper bounding the latter. 
		Under the given setting, $V(\cdot, \cdot \, ; \psi) = \psi(a,a') + \frac{1}{2}(\psi(a',a) - \psi(a,a') ) = \frac{1}{2} (\psi(a',a) + \psi(a',a) )$ for all choices of $\psi$.
		As such, 
		\begin{align}
		\bcM_t^a &\in \argmax_{\sM \in S(V(\cdot, \cdot \, ; \mu ) ) } \bbE[U_t(a, \sM(a) ; \nu_t)] \nonumber
		\\
		&= \argmax_{\sM \in S(V(\cdot, \cdot \, ; \mu ) )}  \frac{1}{2} ( \mu(a, \sM(a)) +  \mu(\sM(a), a) )  \nonumber
		\\
		&= \argmax_{\sM \in S(\rho)}  \frac{1}{2} \rho(a,\sM(a)) , \label{eq:transfer_no_cost_eq2}
		\end{align}
		where $\rho(a,a') = \frac{1}{2}(\mu(a,a') + \mu(a', a))$,
		and
		the second equality follows from \eqref{eq:transfer_no_cost_eq1}.
		By assumption (see Section \ref{sec:setup}), $\rho$ is pairwise-unique. 
		Applying Lemma \ref{lem:pairwise_unique_matching}, $| S(\rho) | = 1$. In  other words, there is a unique stable matching under $\rho$. 
		Let $\sM^{*,\rho}$ denote this stable matching. 
		Since both optimal and pessimal regret are from the same set $S(\rho)$ and $|S(\rho)| = 1$, $\bcM_t^a = \ucM_t^a =  \sM^{*,\rho}$, and $\uR(a ; \ccM) = \bR(a, \ccM)$ for all $a \in \cA$ and all $t \in [T]$.

		By \eqref{eq:transfer_no_cost_eq1}, 
		\begin{align}
		\bR(a ; \ccM) &= \sum_{t=1}^T \bbE\left[ U_t(a, \bcM_t^a(a) ; \nu_t) - U_t(a , \ccM(a ; \nu_t); \nu_t) \right] \nonumber
		\\
		&= \sum_{t=1}^T 
		 \frac{1}{2} \bbE[ \mu(a,\bcM_t^a(a)) +  \mu(\bcM_t^a(a),a) 
				- \mu(a, \ccM(a ; \nu_t )) -  \mu(\ccM(a ; \nu_t ),a) ] \nonumber
		 \\
		 &= \sum_{t=1}^T 
		 \frac{1}{2} \left(  \mu(a, \sM^{*,\rho}(a)) +  \mu( \sM^{*,\rho}(a),a) 
		 -  \bbE[ \mu(a, \ccM(a ; \nu_t )) +  \mu(\ccM(a ; \nu_t ),a) ] \right) \nonumber
		 \\
		 &= \sum_{t=1}^T 
		 \frac{1}{2} \left(  \rho(a, \sM^{*,\rho}(a)) 
		 -  \bbE[ \rho(a, \ccM(a ; \nu_t ))  ] \right) \nonumber
		  \\
		 &=  \frac{1}{2}  \sum_{\sM \in \cW} \bbE[T( \ccM( \cdot \, ; \nu_t) = \sM(\cdot))]
		\left(  \rho(a, \sM^{*,\rho}(a)) 
		 -   \rho(a, \sM(a ))  \right) \nonumber
		 \\
		 &\leq  \frac{1}{2}  \sum_{\sM \in \cW \setminus \sM^{*,\rho}} \bbE[T( \ccM( \cdot \, ; \nu_t) = \sM(\cdot))]
		 \Delta^*(a, \sM(a )) \nonumber
		  \\
		 &\leq  \frac{1}{2} \max_{a' \in \cA^+} \Delta^*(a, a')   \bbE[T( \ccM( \cdot \, ; \nu_t) \neq \sM^{*,\rho} (\cdot) )] ,		 
		  \label{eq:transfer_no_cost_eq3}
		\end{align}
		where $\Delta^*(a,a') = \rho(a,\sM^{*,\rho}(a)) - \rho(a,a')$.
		Similarly to the proof in the no-cost, no-transfer setting, 
		we would like to apply Lemma \ref{lem:no_cost_transfer_T_bound}, but we first need to transform the current setting into an equivalent setting that meets the conditions of Lemma \ref{lem:no_cost_transfer_T_bound}. 
		
		To do so, consider a new setting $(\cA, \mu')$, in which there is no cost or transfer.
		In this new setting, let the true and transient preferences at time $t$ be denoted by $\mu'$, where $\mu'(a,a') = \frac{1}{2} (\mu(a,a') + \mu(a',a))$. 
		All other quantities are defined analogously with respect to $\mu'$ (e.g., $\nu_t'$ is defined as in Section \ref{sec:setup} with respect to $\mu'$, $X_t'(a,a')$ are drawn i.i.d. from a $\sigma^2$-sub-Gaussian distribution centered at $\mu'(a,a')$).
		Furthermore, let $\Delta^{\rho}_{\min } = \min_{a_1,a_2,a_3 \in \cA} |\mu' (a_1, a_2) - \mu'(a_1, a_3)|$.
		This is indeed an equivalent setting to that stated in Lemma \ref{lem:no_cost_transfer_T_bound}, except that $U_t' \neq X'_t$. 
		However, as evident in the proof of Lemma \ref{lem:no_cost_transfer_T_bound}, $U_t'$ is not relevant to the proof. Moreover, recall that agents observe their reward $X_t$ at each time step and know how much is transferred. Agents can therefore distinguish between their reward and transfer. Combining these two facts, Lemma \ref{lem:no_cost_transfer_T_bound} can be applied because all relevant quantities meet the lemma's conditions.
		In particular, the only two quantities of interest are $V(a , a' ; \mu) = \frac{1}{2}(\mu(a,a') + \mu(a',a)) = V'(a,a'; \mu')$ and $V(a , a' ; \nu_t) = \frac{1}{2}(\nu_t(a,a') + \nu_t(a',a)) = V'(a,a'; \nu_t')$, as required. 
		
		Then, applying Lemma \ref{lem:no_cost_transfer_T_bound} to the equivalent alternate setting $(A, \mu')$:  
		\begin{align}
		\bbE T'(\sM(\cdot \, ; \nu'_t) \notin S(\mu')  ) = \bbE[T( \ccM( \cdot \, ; \nu_t) \neq \sM^{*,\rho} (\cdot) )]
		 &\leq 2 N^2 L    \left( \frac{8 \sigma^2 \alpha \log(T)}{ (\Delta^{\rho}_{\min})^2} +  \frac{\alpha}{\alpha - 2} \right) . \label{eq:transfer_no_cost_eq4}
		\end{align} 
		By definition of $\mu'$, $\Delta^{\rho}_{\min} = \frac{1}{2}\min_{a_1 \in \cA,a_2 \in \cA^+, a_3 \in \cA^+ \setminus a_2} |\mu(a_1,a_2) + \mu(a_2,a_1) -  \mu(a_1,a_3) - \mu(a_3,a_1)|$.
		Combining \eqref{eq:transfer_no_cost_eq4} with \eqref{eq:transfer_no_cost_eq3} yields the result on optimal regret. 
		
		\noindent \textbf{Fairness is guaranteed}. 
		That $\ccM$ as described in the result also ensures fairness (i.e., stability guarantees fairness)  follows directly from the analysis under ``Logarithmic optimal regret'' above. 
		Specifically, since $\bR(a ; \ccM) = O(\log(T))$ for all $a \in \cA$, fairness is achieved for all agents by Definition \ref{def:fairness}. 

		\noindent \textbf{High social welfare is guaranteed}. 
		That $\ccM$ also results in high social welfare (i.e., stability guarantees high social welfare)  follows directly from 
		the fact that the greedy algorithm in Algorithm \ref{alg:greedy_balanced} returns a matching whose summed edge weights is within factor-2 of the maximal bipartite weighted matching. 
		
		Recall that the social welfare of $(\ccM, \cC, \cT)$ at time $t$ is defined as $W_t(\ccM) = \sum_{a \in \cA} V(a, \ccM(a ; \nu_t) ;  \nu_t)$.
		Under \eqref{eq:T_1}, $V(a,a' ; \nu_t) = \frac{1}{2} (\nu_t(a,a') + \nu_t(a',a))$.  
		Recall from above that applying Algorithm \ref{alg:greedy_balanced} to preferences $\nu_t$ returns a stable matching under payoffs $V(\cdot, \cdot \, ; \nu_t)$. 
		It remains to show that the social welfare obtained 
		by applying Algorithm \ref{alg:greedy_balanced} to preferences $\nu_t$ is within factor-2 of maximum social welfare possible under any feasible (not necessarily stable) matching in $\cW$ for all $t \in [T]$. 
		
		Consider Algorithm \ref{alg:greedy_balanced}. 
		In the remainder of the proof, $e = (u,p)$, $e' = (u', p')$, and $e'' = (u'', p'')$. 
		If an edge $e = (u,p)$ is added by the algorithm, then it does not conflict with any of the edges added before it. 
		In other words, there are no edges $e' = (u', p')$ for which $w(e') > w(e)$ such that $u' = u$ or $p' = p$. 
		Applying this observation, we find that,
		for any $\sM \in \cW$,
		\begin{align*}
		 \sum_{a \in \cA} V(a, \sM(a) ; \nu_t) 
		&= \sum_{a \in \cA} \frac{1}{2}(\nu_t(a, \sM(a)) + \nu_t( \sM(a), a )) 
		\\
		&= \sum_{e  : \sM(u) = p} (\nu_t(u, p) + \nu_t(p, u))
		\\
		&= \sum_{e  : \sM(u) = p} w(e)
		\\
		&\leq \sum_{e : \sM(u) = p} \max_{e' \neq e'' : u' = u \cap p'' = p} (w(e') + w(e''))
		\\
		&\leq 
		\sum_{e : \sM(u) = p} \left( \max_{e' : u' = u }  w(e') + \max_{e'' : p'' = p}  w(e'') \right)
		\\
		&\leq 
		\sum_{e : \sM(u) = p} \left(  w((u, \ccM(u ; \nu_t) )) + w((\ccM(p ; \nu_t), p)) \right)
		\\
		&\leq \sum_{u \in \cU}  w((u, \ccM(u ; \nu_t) )) + \sum_{p \in \cP} w((\ccM(p ; \nu_t), p)) 
	\end{align*}
\begin{align*}
		&= \sum_{a \in \cA}  \nu_t(a , \ccM(a ; \nu_t )) + \nu_t(\ccM(a ; \nu_t ), a)
		\\
		&= 2 \sum_{a \in \cA} V(a , \ccM(a ; \nu_t ) ; \nu_t)
		\\
		&= 2 W_t(\ccM) .
		\end{align*}
		Since $\sum_{a \in \cA} V(a, \sM(a) ; \nu_t)  \leq 2  W_t(\ccM)$ for all $\sM \in \cW$, $\max_{\sM \in \cW} \sum_{a \in \cA} V(a, \sM(a) ; \nu_t)  \leq 2 W_t(\ccM)$ for all $t \in [T]$, which concludes the proof. 		
\end{proof}

\section{PROOFS FOR PRICING SETTING (SECTION \ref{sec:MP})} \label{sec:app_pricing}

	\begin{lemmaApp} \label{lem:MP_rule_unique}
		Suppose that $\cC$ and $\cT$ are set according to \eqref{eq:M_1}-\eqref{eq:M_3}. 
		Suppose there exists a $B \in \bbR_{> 0}$ such that $ | \mu(u, \cdot) | \leq B$ for all $u \in \cU$. 
		For any arbitrary ordering of providers $(p_1, p_2, \hdots, p_L)$, let $c_1 = 0$, $c_2 = 2 B(1 - L)$, and $g(p_k ; \psi) = 2 B (L - k)$ for all $p_k \in \cP$ and $\psi: \cA \times \cA^+ \rightarrow \bbR$. 
		Then, $|S(V(\cdot, \cdot \, ; \mu)) | = 1$, i.e., there is a unique stable matching under preferences $\mu$. 
	\end{lemmaApp}
	\begin{proof}
		By Lemma \ref{lem:unique_matching_proposer_opt}, there is a unique stable matching under payoffs $V: \cA \times \cA^+ \rightarrow \bbR_{\geq 0}$ where $a' \neq a'' \implies V(a,a') \neq V(a,a'') \forall a \in \cA$ if performing the GS algorithm with providers as proposers and with users as proposers return the same matching $\sM^{*,{\scriptscriptstyle B}}$.
		We now show that this is indeed the case for $c_1 = 0$, $c_2 = 2B(1 - L)$, and $g(p_k ; \psi) = 2B(L - k)$. 
		
		Recall that the GS algorithm with providers and users as proposers can be found in Algorithms \ref{alg:GS} and \ref{alg:GS_user}, respectively. 
		Suppose that $\sM^{*,{\scriptscriptstyle B}}$ is returned by Algorithm \ref{alg:GS}. 
		Recall that $\sM^{*,{\scriptscriptstyle B}}$ is agnostic to which proposer is chosen in Line \ref{line:pick_proposer} as long as the proposer is unmatched. 
		Similarly, Algorithm \ref{alg:GS_user} is agnostic to which user is chosen in Line \ref{line:GS_pick_user} as long as the user is unmatched and as a non-empty queue. 
		To show that Algorithm \ref{alg:GS_user} returns $\sM^{*,{\scriptscriptstyle B}}$, we show that there is a way to pick 
		the user that is proposing in Line \ref{line:GS_pick_user} such that the resulting matching is $\sM^{*,{\scriptscriptstyle B}}$. 
		
		First, note that for any user $u \in \cU$, we have that: 
		$V(u , p_k ; \mu) = \mu(u,p_k) - g(p_k ; \mu ) = \mu(u, p_k) - 2 B (L - k)$, which implies that $V(u, p_{k} ; \mu) - V(u, p_{k-1} ; \mu ) = \mu(u, p_{k}) - \mu(u, p_{k-1}) - 2 B (L - k)  + 2  B (L - k + 1) \geq - B + 2 B = B > 0$.  By induction, $V(u, p_k ; \mu) > V(u, p_j ; \mu)$ for all $k > j$ and  all $u \in \cU$. That is, all users have the same ordinal preferences over providers: $p_L \succ_{V(u, \cdot \, ; \mu)} p_{L-1} \succ_{V(u, \cdot \, ; \mu)} \cdots \succ_{V(u, \cdot \, ; \mu)} p_1$ for all $u \in \cU$. 
		
		We begin with Algorithm \ref{alg:GS_user}. 
		Let user $\sM^{*,{\scriptscriptstyle B}}(p_L)$ be the first to propose.
		$\sM^{*,{\scriptscriptstyle B}}(p_L)$ proposes to $p_L$ since $p_L$ is the top preference of all users. 
		$p_L$ accepts since it is unmatched. 
		Then, let $\sM^{*,{\scriptscriptstyle B}}(p_{L-1})$ propose twice. 
		It first proposes to $p_L$ and is rejected (if not, then $p_L$ prefers $\sM^{*,{\scriptscriptstyle B}}(p_{L-1})$ over $\sM^{*,{\scriptscriptstyle B}}(p_L)$, and we already know that all users prefer $p_L$ the most, which means that $\sM^{*,{\scriptscriptstyle B}}$ is not stable, which is a contradiction). 
		It then proposes to $p_{L-1}$ who accepts since it is unmatched. 
		If we proceed in such a manner, we eventually get to a point in the while loop of Algorithm \ref{alg:GS_user} at which all users and providers are matched according to $\sM^{*,{\scriptscriptstyle B}}$. 
		However, the while loop has not terminated. 
		Even so, the matching will not change because, if a user $u$ proposes after this point in the while loop,  the $p$ to which $u$ proposes must be matched. 
		$u$ therefore prefers to be matched to $p$ than to be unmatched, but $p$ cannot prefer to be matched to $u$ than to $\sM^{*,{\scriptscriptstyle B}}(p)$ because such a result would imply that $\sM^{*,{\scriptscriptstyle B}}$ is not stable, which is a contradiction. Therefore, the while loop continues until termination, and  Algorithm \ref{alg:GS_user}  returns $\sM^{*,{\scriptscriptstyle B}}$. 
		Since both Algorithm \ref{alg:GS} and \ref{alg:GS_user} return the same matching $\sM^{*,{\scriptscriptstyle B}}$, $\sM^{*,{\scriptscriptstyle B}}$ is the unique stable matching by Lemma \ref{lem:unique_matching_proposer_opt}.
	\end{proof}

	\begin{lemmaApp}
		Suppose that $\cC$ and $\cT$ are set according to \eqref{eq:M_1}-\eqref{eq:M_3}. 
		Suppose there exists a $B \in \bbR_{> 0}$ such that $ | \mu(u, \cdot) | \leq B $ for all $u \in \cU$. 
		Then, there always exist constants $c_1, c_2$ and pricing $g$ such that  $|S(V(\cdot, \cdot \, ; \mu)) | = 1$, i.e., there is a unique stable matching under payoffs $V(\cdot, \cdot \, ; \mu)$. 
	\end{lemmaApp}
	\begin{proof}
		This result follows directly from Lemma \ref{lem:MP_rule_unique}, which gives one set of costs $c_1, c_2$  and pricing rule $g$ under which $|S(V(\cdot, \cdot \, ; \mu)) | = 1$.
	\end{proof}

\begin{theorem}
	Suppose that $\cC$ and $\cT$ are set according to \eqref{eq:M_1}-\eqref{eq:M_3}. 
	If the GS algorithm is applied over $V(\cdot, \cdot \, ; \nu_t)$ at every $t \in [T]$, then the system is stable. 
	Moreover, if  there exists a $B \in \bbR_{> 0}$ such that $| \mu(u, \cdot) | \leq B$ for all $u \in \cU$,
	then 
	$| S(V(\cdot, \cdot \, ; \mu)) | = 1$
	and
	there exist constants $c_1$ and $c_2$ as well as a pricing rule $g$ such that
	$\ccM$ is fair
	and 
	$\uR (a; \ccM) = \bR(a ; \ccM) \leq 
	2 \Delta_{\max}^{*,{\scriptscriptstyle B}}(a) N^2 L \left( \frac{8 \sigma^2 \alpha \log(T)}{ (\Delta_{\min} )^2} +  \frac{\alpha}{\alpha - 2} \right)$
	for all $a \in \cA$, 
	where
	$\sM^{*,{\scriptscriptstyle B}}$ is the only element in $S(V(\cdot, \cdot \, ; \mu))$
	and $\Delta^{*,{\scriptscriptstyle B}}_{\max}(a) = 2 B(L - 1) \mathbf{1}(a \in \cU) + \max_{a' \in \cA^+} ( \mu(a, \sM^{*,{\scriptscriptstyle B}}(a) )  - \mu(a, a' ))$.
\end{theorem}
	\begin{proof}
		By Lemma \ref{lem:GS_defection_proof}, the system is stable since the GS algorithm is applied at each time step. 
		Next, we perform the regret analysis. 
		
		\noindent \textbf{Logarithmic optimal regret}.
		By \eqref{eq:M_1}-\eqref{eq:M_3}, $\cC(p,u ; \nu_t) = c_1$, $\cC(u,p ; \nu_t) = c_2$,
		$\cT(p, u ; \nu_t)  = g(p; \nu_t)$, 
		and $\cT(u, p ; \nu_t)  = -g(p ; \nu_t)$ 
		for all $u \in \cU$, $p \in \cP$, and $t \in [T]$.
		As such, the observed payoffs at each time step $t$ are 
		$U_t(p, u ; \nu_t) = X_t(p, u) - c_1 + g(p  ; \nu_t)$
		and $U_t(u, p ; \nu_t) = X_t(u, p) - c_2 - g(p  ; \nu_t)$ for all $u \in \cU$ and $p \in \cP$. 
		In addition,
		$\bbE[U_t(p , u ; \nu_t)] 
		= \mu(p,u)  -c_1 + \bbE[g(p ; \nu_t)]$, 
		$\bbE[U_t(u , p ; \nu_t)] 
		= \mu(u,p)  - c_2 - \bbE[g(p ; \nu_t)]$,
		$V(p, u; \psi) = \psi(p,u) - c_1 + g(p ; \psi)$, 
		and 
		$V(u, p ; \psi) = \psi(u,p) - c_2 - g(p ; \psi)$. 
		
		Recall that our goal is to show that there exists a pricing rule $g$ such that both pessimal and optimal regret are upper bounded by $O(\log(T))$. 
		To show existence, let $c_1 = 0$, $c_2 = 2 B (1 - L)$,
		$g(p_k; \psi) = 2 B (L - k) $ for all $k \in [L]$ and preferences $\psi$. 
		By Lemma \ref{lem:MP_rule_unique}, $|S(V(\cdot, \cdot \, ; \mu))| = 1$, which implies that $\bcM_t^a = \ucM_t^a$. We denote this unique stable matching under payoffs $V(\cdot, \cdot \, ; \mu)$ by $\sM^{*,{\scriptscriptstyle B}}$. 
		Since $|S(V(\cdot, \cdot \, ; \mu))| = 1$, $\ushort{R}(a ; \ccM) = \bar{R}(a ; \ccM)$. 
		We focus on upper bounding the latter.
		For any provider $p_k \in \cP$,
		\begin{align}
		\bar{R}(p ; \ccM) &= 
		 \sum_{t=1}^T 
		(\mu(p, \sM^{*,{\scriptscriptstyle B}}(p) ) - c_1 + \bbE[g(\sM^{*,{\scriptscriptstyle B}}(p)  ; \nu_t)] - \mu(p, \ccM(p; \nu_t ))  + c_1 - \bbE[g(\ccM(p; \nu_t ) ; \nu_t)] )  \nonumber
		\\
		&= \sum_{t=1}^T 
		(\mu(p, \sM^{*,{\scriptscriptstyle B}}(p) ) - \mu(p, \ccM(p; \nu_t )) + 2 B (L - k) - 2 B (L - k) ) \nonumber
		\\
		&= \sum_{\sM \in \cW} 
		\bbE T(\ccM( \cdot \, ; \nu_t) = \sM(\cdot) ) (\mu(p, \sM^{*,{\scriptscriptstyle B}}(p) ) - \mu(p, \sM (p)) ) \nonumber
		\\
		&\leq \Delta_{\max}^{*,{\scriptscriptstyle B}}(p) \sum_{\sM \neq \sM^{*,{\scriptscriptstyle B}}} 
		\bbE T(\ccM( \cdot \, ; \nu_t) = \sM(\cdot) )  \nonumber 
		\\
		&\leq \Delta_{\max}^{*,{\scriptscriptstyle B}}(p) 
		\bbE T(\ccM( \cdot \, ; \nu_t) \neq \sM^{*,{\scriptscriptstyle B}}(\cdot) )  \label{eq:market_price_eq1} ,
		\end{align}
		where $\Delta_{\max}^{*,{\scriptscriptstyle B}}(p) =  \max_{u \in \cU} (\mu(p,\sM^{*,{\scriptscriptstyle B}}(p)) - \mu(p,u))$. 
		For all users $u \in \cU$,
		\begin{align}
		\bar{R}(u ; \ccM) &= 
		\sum_{t=1}^T 
		(\mu(u, \sM^{*,{\scriptscriptstyle B}}(u) )  - c_2 - \bbE[g(\sM^{*,{\scriptscriptstyle B}}(u)  ; \nu_t)] - \mu(u, \ccM(u ; \nu_t )) + c_2 + \bbE[g(\ccM(u ; \nu_t ) ; \nu_t)] ) \nonumber
		\\ 
		&=
		\sum_{t=1}^T 
		(\mu(u, \sM^{*,{\scriptscriptstyle B}}(u) )  - \bbE[g(\sM^{*,{\scriptscriptstyle B}}(u)  ; \nu_t)] - \mu(u, \ccM(u ; \nu_t ))  + \bbE[g(\ccM(u ; \nu_t ) ; \nu_t)] ) \nonumber
		\\
		&\leq \Delta_{\max}^{*,{\scriptscriptstyle B}}(u)  \sum_{\sM \neq \sM^{*,{\scriptscriptstyle B}}} \bbE T(\ccM(\cdot \, ; \nu_t) = \sM(\cdot) ) \nonumber 
		\\
		&= \Delta_{\max}^{*,{\scriptscriptstyle B}}(u) \bbE T(\ccM(\cdot \, ; \nu_t) \neq \sM^{*,{\scriptscriptstyle B}}(\cdot) ) ,
		\label{eq:market_price_eq2}
		\end{align}
		where $\Delta_{\max}^{*,{\scriptscriptstyle B}}(u) = 2 B(L - 1) + \max_{p \in \cP} (\mu(u, \sM^{*,{\scriptscriptstyle B}}(u) )  - \mu(u, p ))$. 
		
		We therefore would like to upper bound $\bbE T(\ccM( \cdot \, ; \nu_t) \neq \sM^{*,{\scriptscriptstyle B}}(\cdot) )$. 
		Recall that $\ccM( \cdot \, ; \nu_t) = S(V(\cdot, \cdot \, ; \nu_t))$ and $\sM^{*,{\scriptscriptstyle B}} \in S(V(\cdot, \cdot \, ; \mu))$.
		Then, following approximately the same steps as in the proof in Lemma \ref{lem:no_cost_transfer_T_bound}:
		\begin{align}
		\bbE T( \ccM( \cdot \, ; \nu_t)  \neq \sM^{*,{\scriptscriptstyle B}}(\cdot) ) &\leq \bbE T(\exists u', u'' \in \cU, p', p'' \in \cP : \ccM(u' ; \nu_t) = p' \cap  \ccM(u'' ; \nu_t) = p''  \nonumber
		\\
		& \qquad \cap V(u', p'' ; \mu) > V(u', p' ; \mu)  \nonumber
		\\
		& \qquad \cap V(p'', u' ; \mu) > V(p'', u'' ; \mu) \nonumber
		\\
		& \qquad \cap ( V(u', p'' ; \nu_t) \leq V(u', p' ; \nu_t) \cup V(p'', u' ; \nu_t) \leq V(p'', u'' ; \nu_t )  )  ) \nonumber
		\\
		&\leq \bbE T(\exists u', u'' \in \cU, p', p'' \in \cP : \ccM(u' ; \nu_t) = p' \cap  \ccM(u'' ; \nu_t) = p''  \nonumber
		\\
		& \qquad \qquad\cap V(u', p'' ; \mu) > V(u', p' ; \mu)  \nonumber
		\\
		& \qquad \qquad\cap V(p'', u' ; \mu) > V(p'', u'' ; \mu) \nonumber
		\\
		& \qquad \qquad\cap  V(u', p'' ; \nu_t) \leq V(u', p' ; \nu_t)  ) \nonumber
		\\
		&\qquad + \bbE T(\exists u', u'' \in \cU, p', p'' \in \cP : \ccM(u' ; \nu_t) = p' \cap  \ccM(u'' ; \nu_t) = p''  \nonumber
		\\
		& \qquad\qquad \cap V(u', p'' ; \mu) > V(u', p' ; \mu)  \nonumber
		\\
		& \qquad \qquad\cap V(p'', u' ; \mu) > V(p'', u'' ; \mu) \nonumber
		\\
		& \qquad\qquad \cap V(p'', u' ; \nu_t) \leq V(p'', u'' ; \nu_t )  ) \nonumber
		\\
		&\leq \bbE T(\exists u \in \cU, p', p'' \in \cP : \ccM(u ; \nu_t) = p' \nonumber
		\\
		& \qquad \qquad\cap V(u, p'' ; \mu) > V(u, p' ; \mu)  \nonumber
		\\
		& \qquad \qquad\cap  V(u, p'' ; \nu_t) \leq V(u, p' ; \nu_t)  ) \nonumber
		\\
		&\qquad + \bbE T(\exists u', u'' \in \cU, p \in \cP :  \cap  \ccM(u'' ; \nu_t) = p \nonumber
		\\
		& \qquad \qquad\cap V(p, u' ; \mu) > V(p, u'' ; \mu) \nonumber
		\\
		& \qquad\qquad \cap V(p, u' ; \nu_t) \leq V(p, u'' ; \nu_t )  ) . \label{eq:market_price_eq3}
		\end{align}
		Let's split \eqref{eq:market_price_eq3} into two terms. Examining the second term, 
		\begin{align}
		\bbE&T(\exists u', u'' \in \cU, p \in \cP : \ccM(u'' ; \nu_t) = p 
		\cap V(p, u' ; \mu) > V(p, u'' ; \mu) 
		\cap V(p, u' ; \nu_t) \leq V(p, u'' ; \nu_t )  ) \nonumber
		\\
		&=  \bbE T(\exists u', u'' \in \cU, p \in \cP : \ccM(u'' ; \nu_t) = p \nonumber
		\\
		& \qquad\qquad\qquad \cap \mu(p, u')  - c_1 + g(p ; \mu) > \mu(p, u'' ) - c_1 + g(p ; \mu)  \nonumber
		\\
		& \qquad\qquad\qquad\cap  \nu_t(p, u' ) - c_1 + g(p ; \nu_t)  \leq  \nu_t(p, u'') - c_1 + g(p ; \nu_t)  ) \nonumber
		\\
		&=  \bbE T(\exists u', u'' \in \cU, p \in \cP : \ccM(u'' ; \nu_t) = p \cap \mu(p, u') > \mu(p, u'' )   \nonumber
		\cap  \nu_t(p, u' )  \leq  \nu_t(p, u'') )  \nonumber
		\\
		&\leq N^2 L \left( \frac{8 \sigma^2 \alpha \log(T)}{\Delta_{\min}^2}  + \frac{\alpha}{\alpha - 2} \right) , \label{eq:market_price_eq4}
		\end{align}
		where $\Delta_{\min}= \min_{a, a', a'' \in \cA } | \mu(a,a') - \mu(a, a'') |$
		and  the last inequality follows from Lemma \ref{lem:ucb_subgauss} (for further explanation, see the proof of Lemma \ref{lem:no_cost_transfer_T_bound}). 
		
		Turning to the first term of \eqref{eq:market_price_eq3}, 
		\begin{align}
		\bbE&T(\exists u \in \cU, p', p'' \in \cP : \ccM(u ; \nu_t) = p' \cap V(u, p'' ; \mu) > V(u, p' ; \mu) \cap  V(u, p'' ; \nu_t) \leq V(u, p' ; \nu_t)  ) \nonumber
		\\
		&=  \bbE T(\exists u \in \cU, p', p'' \in \cP : \ccM(u ; \nu_t) = p' \cap \mu(u, p'' ; \mu) - c_2 - g(p'' ; \mu) > \mu(u, p' ; \mu) - c_2 - g(p' ; \mu) \nonumber
		\\
		&\qquad \cap  \nu_t(u, p'') - c_2 - g(p'' ; \nu_t) \leq \nu_t(u, p') - c_2 - g(p' ; \nu_t)  ) \nonumber
		\\
		&=  \bbE T(\exists u \in \cU, p', p'' \in \cP : \ccM(u ; \nu_t) = p' \cap \mu(u, p'' ; \mu) - g(p'' ; \mu) > \mu(u, p' ; \mu) - g(p' ; \mu) \nonumber
		\\
		&\qquad \cap  \nu_t(u, p'') - g(p'' ; \mu) \leq \nu_t(u, p') - g(p' ; \mu)  ) \nonumber
		\\
		&\leq N^2 L \left( \frac{8 \sigma^2 \alpha \log(T)}{\Delta_{\min}^2}  + \frac{\alpha}{\alpha - 2} \right) , \label{eq:market_price_eq5}
		\end{align}
		where the last equality follows from the fact that $g(\cdot \, ; \mu) = g(\cdot \, ; \nu_t)$, and the last line follows from Lemma \ref{lem:ucb_subgauss}
		 (for further explanation, see the proof of Lemma \ref{lem:no_cost_transfer_T_bound}). 
		
		Combining \eqref{eq:market_price_eq1}-\eqref{eq:market_price_eq5} yields the regret upper bound as stated in the result. 
		
		\noindent \textbf{Fairness is guaranteed}. 
		That $\ccM$ as described in the result also results in fairness (i.e., stability guarantees fairness)  follows directly from the analysis under ``Logarithmic optimal regret'' above. 
		Specifically, since $\bR(a ; \ccM) = O(\log(T))$ for all $a \in \cA$, fairness is achieved for all agents by Definition \ref{def:fairness}. 
	\end{proof}

\end{document}